\documentclass[letterpaper, 10 pt, journal, twoside]{IEEEtran}

\newcommand{\etal}{\MakeLowercase{\textit{et al.}}}
\newcommand{\atan}{\arctan}

\usepackage{graphicx}
\usepackage{amsmath}
\usepackage{amssymb}
\usepackage{booktabs}
\usepackage{nicefrac}
\usepackage{algorithm}
\usepackage[algo2e]{algorithm2e} 
\usepackage{multirow}
\usepackage{mwe}

\usepackage[pagebackref,breaklinks,colorlinks]{hyperref}

\usepackage{balance}
\usepackage{comment}
\usepackage{fancyhdr}

\usepackage[capitalize]{cleveref}
\crefname{section}{Sec.}{Secs.}
\Crefname{section}{Section}{Sections}
\Crefname{table}{Table}{Tables}
\crefname{table}{Tab.}{Tabs.}

\newcommand{\modified}[1]{\color{black}{{#1 }}\color{black}}

\ifCLASSINFOpdf
\else
\fi
\IEEEoverridecommandlockouts
\begin{document}
%
\title{A Practical Calibration Method for RGB Micro-Grid Polarimetric Cameras}

%
%
%

\author{Joaquin Rodriguez, Lew Lew-Yan-Voon,
Renato Martins, and Olivier Morel 
\thanks{Manuscript received: February 24, 2022; Revised April 29, 2022; Accepted July 10, 2022.} 
\thanks{This paper was recommended for publication by Editor Lucia Pallottino upon evaluation of the Associate Editor and Reviewers' comments.}
\thanks{The authors are with ImViA Lab., Université Bourgogne-Franche-Comté, France. {\tt\small
\{joaquin-jorge.rodriguez, lew.lew-yan-voon, renato.martins, olivier.morel\}@u-bourgogne.fr}}
\thanks{This work was supported by the ICUB project funded by the French National Research Agency ANR-17-CE22-0011.} 
\thanks{Digital Object Identifier (DOI): see top of this page.}}

\markboth{IEEE Robotics and Automation Letters. Preprint Version. Accepted July, 2022}
{Rodriguez \MakeLowercase{\textit{et al.}}: A Practical Calibration Method for Polarimetric Cameras}
\nocite{*}
\maketitle


\thispagestyle{fancy}
\fancyhf{} \chead{{This is a preprint version of the paper to appear at IEEE Robotics and Automation Letters (RAL). The final journal version will be available at \url{https://doi.org/10.1109/LRA.2022.3192655}}}
\cfoot{1}

\begin{abstract}
   Polarimetric imaging has been applied in a growing number of applications
    in robotic vision (ex. underwater navigation, glare removal, de-hazing,
    object classification, and depth estimation). One can find on the market
    RGB Polarization cameras that can capture both color and polarimetric
    state of the light in a single snapshot. Due to the sensor's characteristic
    dispersion, and the use of lenses, it is crucial to calibrate these types
    of cameras so as to obtain correct polarization measurements. The calibration
    methods that have been developed so far are either not adapted to this type
    of cameras, or they require complex equipment and time consuming experiments
    in strict setups. In this paper, we propose a new method to overcome the
    need for complex optical systems to efficiently calibrate these cameras.
    We show that the proposed calibration method has several advantages such as that
    any user can easily calibrate the camera using a uniform, linearly polarized
    light source without any a priori knowledge of its polarization state, and
    with a limited number of acquisitions. We will make our calibration code
    publicly available.
\end{abstract}

\begin{IEEEkeywords}
Calibration and Identification, Polarization, Polarimetric Cameras, Degree of Polarization, Multi-Modality.
\end{IEEEkeywords}

\section{Introduction}
\label{sec:intro}
\begin{figure}
\begin{centering}
\begin{tabular}{cc}
\includegraphics[width=0.37\linewidth]{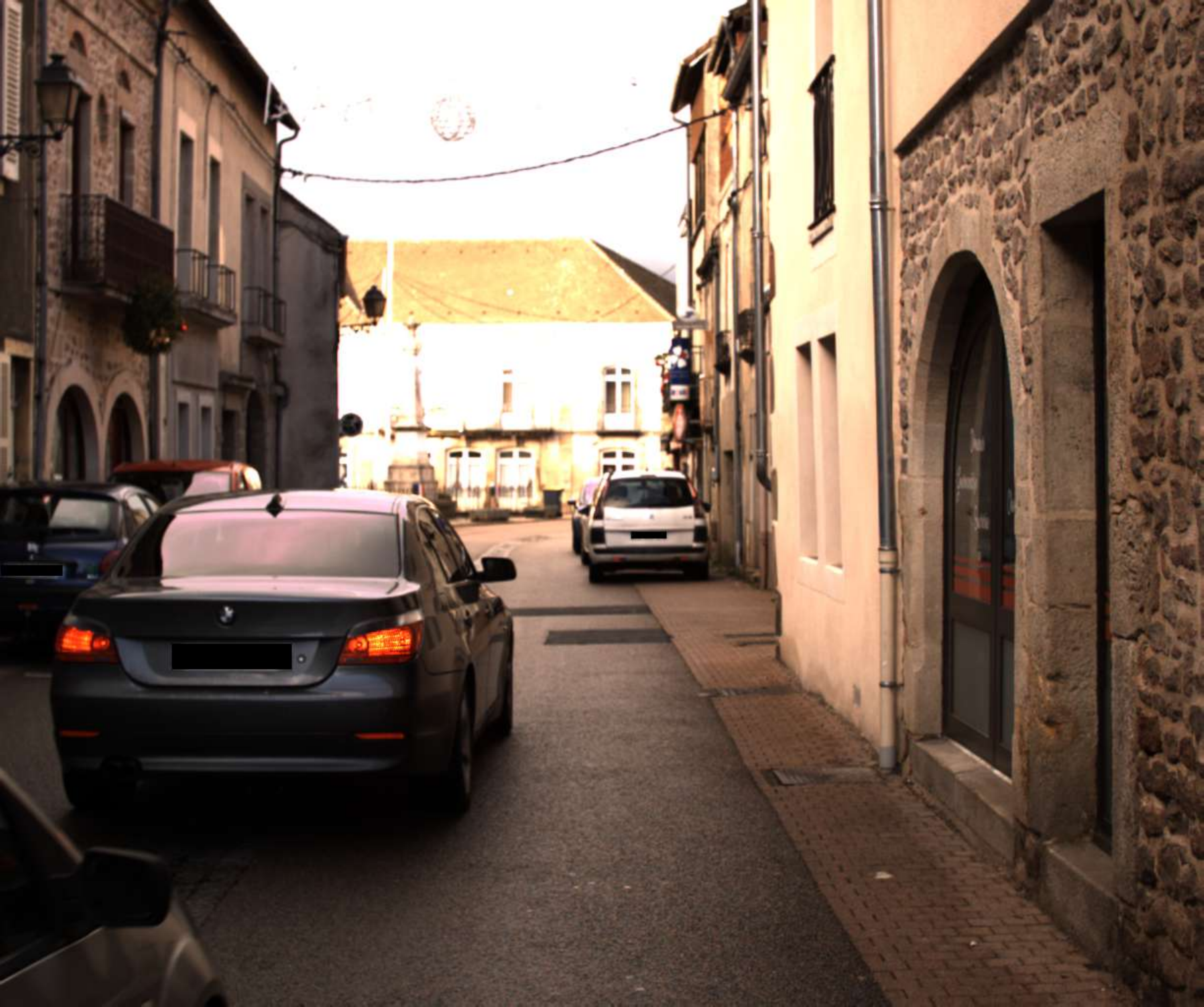} &
\includegraphics[width=0.37\linewidth, height=2.9cm]{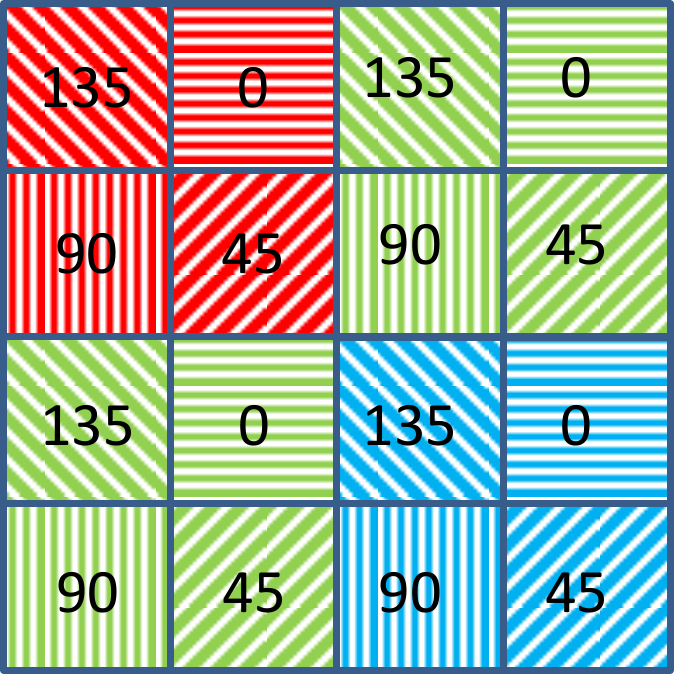} \\
(a) & (b) \\
\includegraphics[width=0.37\linewidth]{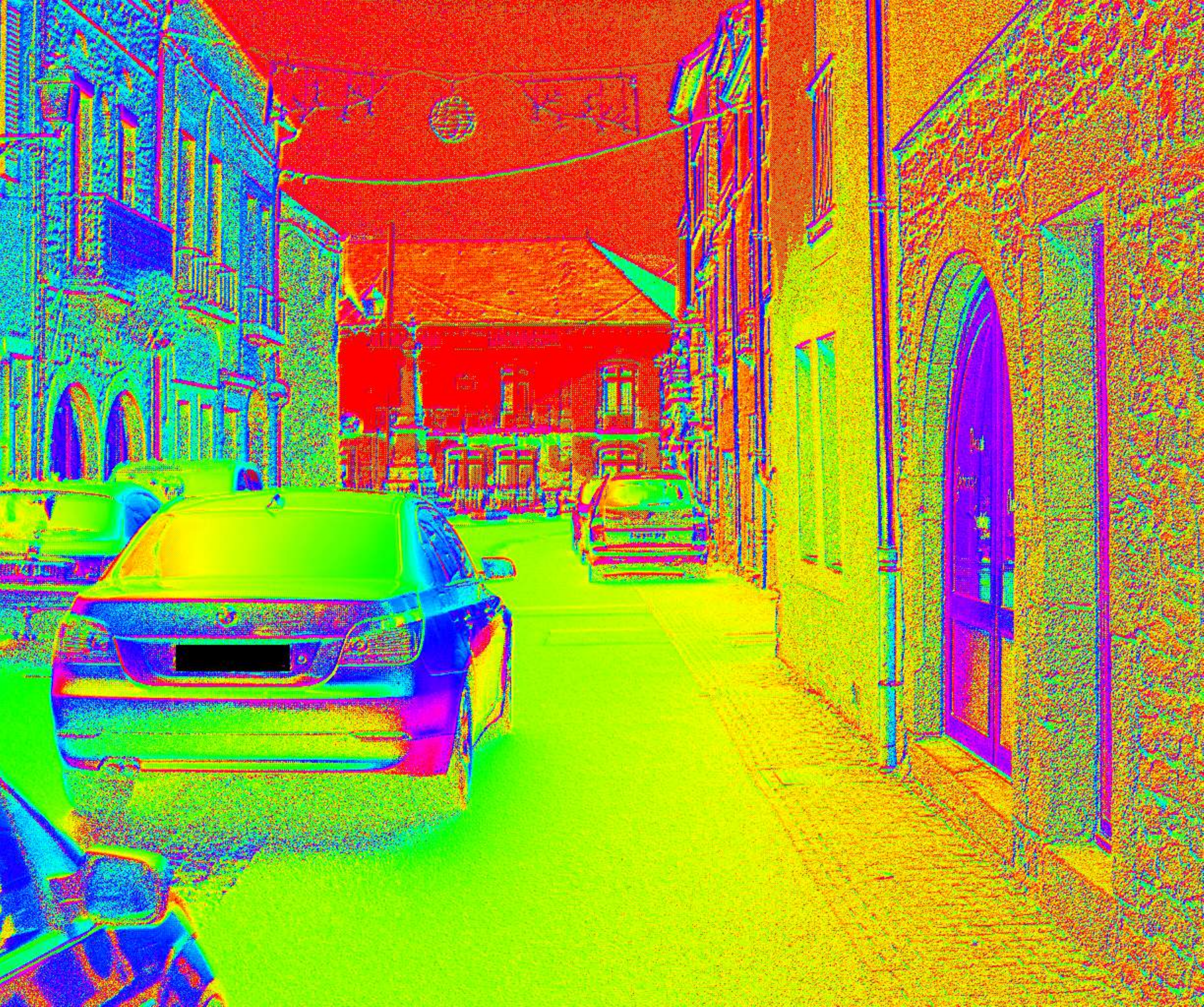} &
\includegraphics[width=0.37\linewidth]{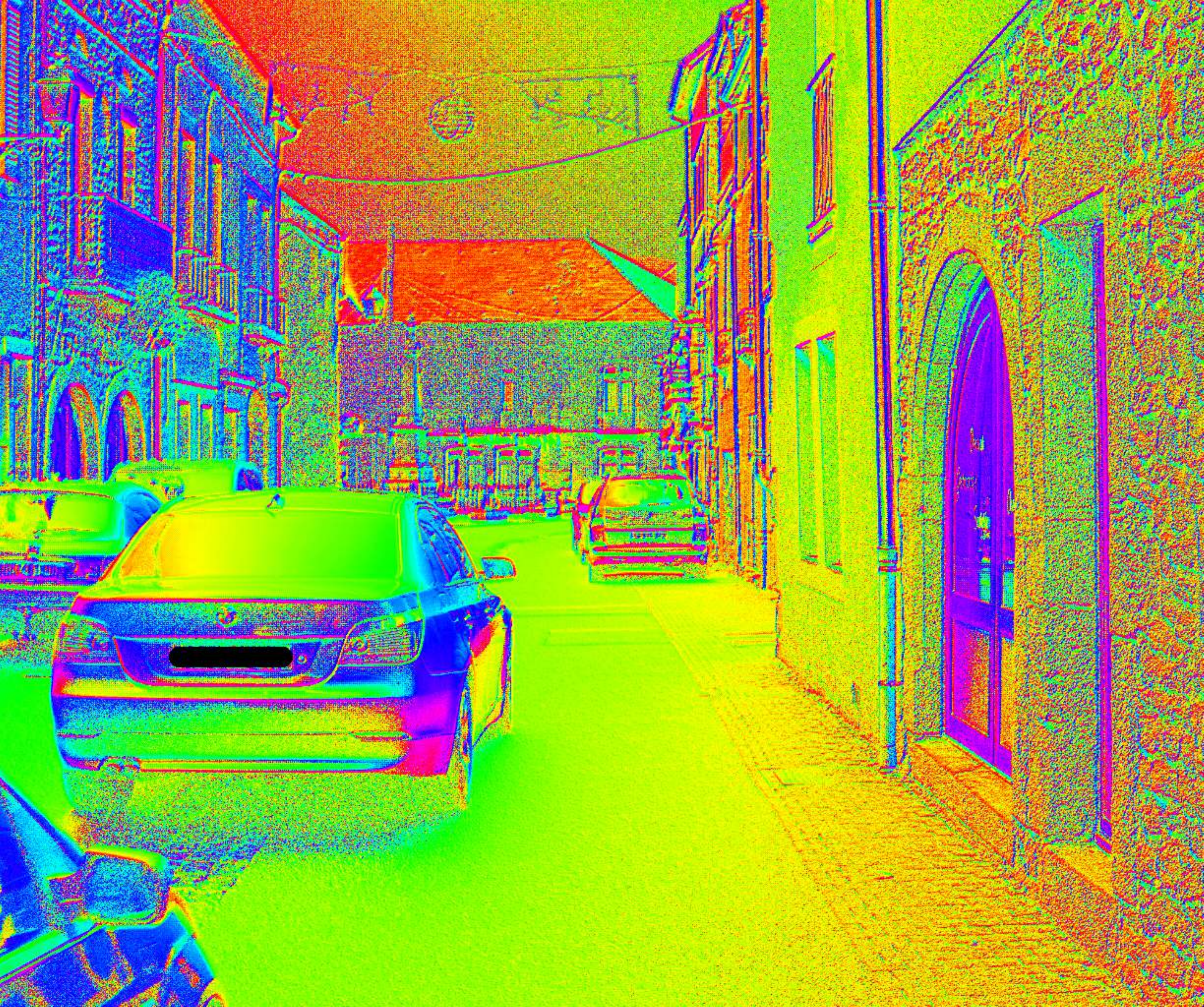} \\
(c) & (d) \\
\end{tabular}
\par\end{centering}\vspace*{-0.2cm}
\caption{\scriptsize\textbf{Effect of the calibration for measuring the angle
of polarization in a real urban scene.} (a) Input color image of the scene captured
with the RGB polarization camera. (b) The four colored super-pixels of the adopted
camera (Sony Polarsens IMX250MYR), arranged following the Bayer pattern
{\modified{with polarizer orientations of $0^\circ$, $45^\circ$, $90^\circ$, and $135^\circ$}}.
(c) Angle of linear polarization image measured with the uncalibrated sensor.
(d) Angle of linear polarization image after calibrating the camera with our
method. Please notice the differences between (c) and (d), where the angle of
polarization of the sky, the building's roof and front wall in the background
were wrongly measured. Without calibration, most of these regions have
the same angle of polarization (identified by the same ``red" color).}
\label{fig:ApplicationExample}\vspace*{-0.7cm}
\end{figure}

\IEEEPARstart{I}{n} recent years, polarization has started to have more interest in the robotics and
computer vision fields \modified{\cite{dsfp, cromo, polaPosePredict, DepthFromStereoPol,
RelPosMarcPo, 9_surfacereconstruction, Ichikawa_2021_CVPR, 6_Solar_spectropolarimetry,
10_3d_reconstruction, 8_water_robotics, 7_skin_cancer, 4_image_dehazing,
12_Overview_of_polarimetric_applications, 11_material_classification, HDRReconsWu,
5_remote_sensing, 28_rt_dense_slam}}. The polarization state of the light is modeled
by the Stokes vector with four components that represent the total incoming light
intensity and the amount of light that is linearly and circularly polarized. Several
types of sensors can be used to measure the polarization state
of the light. Among them, the division of focal plane (DoFP) is the most popular and
the most suitable for real-time applications since it can measure all the required
information to compute the first three Stokes parameters in a single snapshot. It is
composed of super-pixels, that are groups of $2\times2$ pixels, over which are four
directional polarizers at an angle of 0$^\circ$, 45$^\circ$, 90$^\circ$ and
135$^\circ$. \modified{Sony released a DoFP sensor in 2018, the Polarsens IMX250MYR,
that can also measure color information.} It is basically
an array of four colored super-pixels arranged following a Bayer pattern as
represented in \cref{fig:ApplicationExample} (b).

In the Sony Polarsens sensor, the  micro-grid polarizers are formed and
placed under the layer of micro-lenses used to correctly focus the incoming
light onto the pixel, contrary to conventional polarization sensors where it
is above the micro-lenses. In this way, the cross-talking effect due to large
angle of incidence rays is considerably reduced. Nevertheless, the polarization state
determined from the measurement of the light intensities is still inaccurate,
as one can notice in the measured angle of polarization of the real urban
scene shown in \cref{fig:ApplicationExample}. This is because of the sensor
characteristics dispersion due to manufacturing imperfections and to the presence
of the lens in front of the camera. The purpose of calibration is to individually
correct the outputs of the pixels so that they will all have the same value when
the sensor is illuminated with a uniform incident light. Several works attempted
to address this calibration problem
\cite{21_Chen_calib_with_inter_fourier,17_Simil_calib,18_Powell_Gruev}, but the
complex experiments for  generating the calibration data are time consuming,
and require strict experimental setups. In this paper, we describe a practical
method to calibrate cameras based on DoFP RGB polarization sensors. It requires
only a few samples of a uniform and linearly polarized (ULP) light with moderate
values in all of the three RGB channels. The different samples are obtained either
by rotating the camera in front of a ULP light source, or by turning a linear
polarization filter placed between the camera and an unpolarized light source.
Our method will be compared with the well-established super-pixel algorithm, and
we show that, although the experimental setup is simpler, our method exhibits
similar accuracy. The images of degree of polarization, and angle of
polarization will be shown before and after the calibration, concluding that the
calibrated pixel's response is the same over all the sensor area when they are
illuminated with a ULP light.

\section{Related work}
Diverse calibration methods have been developed and reported in the literature to
correct polarization measurements. However, they are either not suitable for a camera
based on an RGB polarization sensor or they require complex equipment, making it
hard to replicate the experiments. For example, the method developed by Schechner
\cite{15_Self_calib_pola} considers a conventional camera with a polarizer filter
in front of it. In this setup, all the pixels share the same polarization filter
and thus, to solve the calibration problem, only a few polarized points in a
generic scene are needed. This is not the case for a DoFP sensor where each
polarization analyzer is composed of four different pixels with polarizers oriented
in four different directions. The method by Wang \etal \cite{lcd_calib_wang} uses an
LCD screen to achieve both, polarimetric and geometric calibration of a camera
mounted with a polarization filter. Nevertheless, this method cannot be used in our
case, because to illuminate all the pixels with the LCD screen,
the sensor to the LCD distance must be so short that the pixel pattern of the screen
is captured by the camera. Thus, the light cannot be considered uniform. Regarding
calibration algorithms dedicated to DoFP sensors, Hagen \etal{} \cite{17_Simil_calib}
proposed a method that requires a few samples only, but the angle of polarisation of
those samples must be known accurately. Chen \cite{21_Chen_calib_with_inter_fourier}
introduced a calibration approach that has several constraints in the
experiment setup: the light source is expected to come from an integrating sphere,
and a band-pass filter is added to estimate the missing pixel through a Fourier-based
approach. Moreover, a motorized rotative polarization filter is used, and all the
light parameters should be known beforehand. Powell and Gruev \cite{18_Powell_Gruev}
calibrate monochrome DoFP polarimeters, by two approaches: the single and super-pixel
algorithm. The drawback of their method is that the calibration set up is too
elaborated to be replicated, since the light information should be known beforehand.
In this paper, a calibration procedure for RGB polarization cameras is presented
with the following contributions:
\begin{enumerate}
  \item[-] A simple and practical pipeline to calibrate ready-to-use polarimetric
    cameras with a lens, in a single step.
  \item[-] A method that does not require any prior knowledge of the polarization
    state of the calibration light samples. This information is estimated using
    only a few
  samples of a ULP light with moderate values for each of the three RGB channels.
  \item[-] We demonstrate the effectiveness of the algorithm by showing that
    only five light samples are enough to calibrate the camera with a
    good accuracy.
\end{enumerate}

\section{Pixel model}
In this section, the pixel model is briefly described using the Stokes vector and
Mueller calculus \modified{\cite{StokesFormalism,23_HuangFei}}. The Stokes vector
is a convenient way of representing the polarization state of the light. It is
composed of four components
\modified{$\mathbf{\underline{S}}=\left[S_{0},S_{1},S_{2},S_{3}\right]$} where $S_{0}$
is the total intensity of the incident light, $S_{1}$ the amount of linearly polarized
light in the horizontal and vertical directions, $S_{2}$ the amount of linearly
polarized light in the $\pm45^\circ$ directions, and $S_{3}$ the amount of circularly
polarized light.

The calibration method that we have developed are for cameras that are based on DoFP
type sensors. These sensors can measure only the linear polarization parameters of the
light. So, only linearly polarized light sources or samples will be used to calibrate
them. Hence, only the first three components of the Stokes vector are considered. In
terms of these three components, the degree of linear polarization (DoLP) and the
angle of linear polarization (AoLP), denoted by $\rho$ and $\alpha$, respectively, are
defined as follows:

\begin{equation}
\rho=\dfrac{\sqrt{S_{1}^{2}+S_{2}^{2}}}{S_{0}};\ \modified{ \alpha=\dfrac{1}{2}\arctan\left(\dfrac{S_{2}}{S_{1}}\right).}
\label{eq:PolarizationParams}
\end{equation}

Conversely, representing the Stokes vector components in terms of $\rho$
and $\alpha$ defined in \cref{eq:PolarizationParams} yields:

\modified{
\begin{equation}
\mathbf{\underline{S}}=\left[
\begin{array}{c}
S_{0} \\
S_{1} \\
S_{2} \\
\end{array}
\right]=\left[
\begin{array}{c}
S_{0} \\
S_{0} \rho \cos\left(2\alpha\right) \\
S_{0} \rho \sin\left(2\alpha\right)\\
\end{array}
\right].
\label{eq:NewStokes}
\end{equation}
}

Now, let us consider a linearly polarized light that passes through an optical
element composed of a pixel of the sensor (with its micro-lens and its micro-polarizer
oriented at $\theta^\circ$), and the lens of the camera to calibrate. The relationship
between the Stokes vector of the input light denoted by
\modified{$\mathbf{\underline{S}}=\left[S_{0},S_{1},S_{2}\right]^{T}$}
and the Stokes vector of the output light denoted by
\modified{$\mathbf{\underline{O_{out}}}=\left[O_{0\theta},O_{1\theta},O_{2\theta}\right]^{T}$},
is given by:

\begin{equation}
\left[\begin{array}{c}
O_{0\theta} \\ O_{1\theta} \\ O_{2\theta}
\end{array}\right]
=
\mathbf{M_{u}}~\left[\begin{array}{c}
S_{0} \\ S_{1} \\ S_{2}
\end{array}\right],
\label{eq:Stokes3}
\end{equation}
where $\mathbf{M_{u}}$ is the Mueller matrix that models all the components in the optical
element, i.e., the camera pixel and the lens.

In the output Stokes vector, only the first component \modified{$O_{0\theta}$}, that corresponds to the
intensity of the light, can be measured by the photosensor device in the pixel. Its expression
in terms of the elements of the Mueller matrix $\mathbf{M_u}$ \cite{23_HuangFei} is equal to:
\begin{equation}
\modified{O_{0\theta} = 
\frac{1}{2}\left[
q+r ~~ \left(q-r\right)\cos\text{\ensuremath{\left(2\theta\right)}} ~~ \left(q-r\right)\sin\text{\ensuremath{\left(2\theta\right)}}\right]
\underline{\mathbf{S}}},
\label{eq:MeasuredIntensity}
\end{equation}
where $q$ and $r$ are, respectively, the major and minor light
transmittance of the linear polarizer, and $\theta$ the orientation of the
micro-filter placed over the pixel \cite{2_Non_uniform_light_calibration}. Denoting the
intensity measured by the photosensor as $I_{\theta}$, then $O_{0\theta}$ is equal to
$I_{\theta}-d_{\theta}$ where $d_{\theta}$ is the dark current noise that must be subtracted
from the measured intensity. Also, letting $T=\frac{q-r}{2}$ be the pixel gain, and
$P=\frac{\left(q-r\right)}{\left(q+r\right)}$  a coefficient that represents the non-ideality
of the filter, \cref{eq:MeasuredIntensity} can be rewritten as:
\begin{equation}
\modified{
I_{\theta}=\left[\begin{array}{ccc}
\frac{T}{P} & T\cos\text{\ensuremath{\left(2\theta\right)}} & T\sin\text{\ensuremath{\left(2\theta\right)}}\end{array}\right]
\underline{\mathbf{S}} +
d_{\theta}.
}
\label{eq:muellerMatrix2}
\end{equation}
\modified{In this equation, $d_{\theta}$ is set equal to zero since the dark current
can be neglected as it will be shown in the experiment section. So, a pixel
with a micro-polarizer at an angle of $\theta^\circ$ will be modelled by the following equation:}

\modified{
\begin{equation}
I_{\theta}=\left[\begin{array}{ccc}
\frac{T}{P} & T\cos\text{\ensuremath{\left(2\theta\right)}} & T\sin\text{\ensuremath{\left(2\theta\right)}}\end{array}\right]
\underline{\mathbf{S}},
\label{eq:PixelModel}
\end{equation}
}
where $\left(T,P,\theta\right)$ are the pixel parameters. For the super-pixel, as shown in
\cref{fig:ApplicationExample} (b), the measured intensities by the four directional pixels in
the $0^\circ$, $45^\circ$, $90^\circ$, $135^\circ$ directions can be stacked to give the
following equation:
\begin{equation}
\left[\begin{array}{c}
I_{0} \\ I_{45} \\ I_{90} \\ I_{135}\end{array}\right]=\mathbf{A}
\left[\begin{array}{c}
S_{0} \\ S_{1} \\ S_{2}
\end{array}\right],
\label{eq:realPol-1}
\end{equation}
where $\mathbf{A}$ is the super-pixel matrix, defined as
\begin{equation}
\mathbf{A}=\left[\begin{array}{ccc}
\nicefrac{T_{0}}{P_{0}} & T_{0}\cos\left(2\theta_{0}\right) & T_{0}\sin\text{\ensuremath{\left(2\theta_{0}\right)}}\\
\nicefrac{T_{45}}{P_{45}} & T_{45}\cos\left(2\theta_{45}\right) & T_{45}\sin\text{\ensuremath{\left(2\theta_{45}\right)}} \\
\nicefrac{T_{90}}{P_{90}} & T_{90}\cos\left(2\theta_{90}\right) & T_{90}\sin\text{\ensuremath{\left(2\theta_{90}\right)}}\\
\nicefrac{T_{135}}{P_{135}} & T_{135}\cos\left(2\theta_{135}\right) & T_{135}\sin\text{\ensuremath{\left(2\theta_{135}\right)}}\\
\end{array}\right].
\label{eq:RealMueller}
\end{equation}
In \cref{eq:RealMueller}, $\left(T_{i},P_{i},\theta_{i}\right)$ are the 
parameters of the pixel with micro-filter orientation
$i=\left\{ 0^\circ,45^\circ,90^\circ,135^\circ\right\}$.
\modified{From Mueller calculus theory, ideally, $T_{i}=0.5$ and $P_{i}=1$
for all $i=\left\{0^\circ,45^\circ,90^\circ,135^\circ\right\}$,
and $\theta_{0^\circ}=0^\circ$, $\theta_{45^\circ}=45^\circ$,
$\theta_{90^\circ}=90^\circ$, and $\theta_{135^\circ}=135^\circ$
for the four pixels of the super-pixel. } However, in a real camera with
a lens, the true values will deviate from the ideal case. The purpose of
our calibration method is to estimate
the true value of these parameters for each sensor's super-pixel.

\section{\label{sec:Algorithm}Calibration method}

An overview of the pipeline of the proposed method is sketched in
\cref{fig:CalibrationPipeline}. It is grounded on the super-pixel
calibration method detailed in \cite{22_Yilbert_2020,18_Powell_Gruev},
which is a well-established method in the literature. The originality
of our calibration algorithm is that no information about the input light
polarization state is required. Instead, we propose to estimate them
and use the estimated polarization state in the calibration method.
In \cref{sec:GeneralCalib}, the super-pixel calibration method is detailed,
in which the input light polarization parameters are required. Then, in
\cref{subsec:AoLPEst} and \cref{sec:IntensityEst}, we present methods to
estimate these light parameters.

\subsection{\label{sec:GeneralCalib}Description of the method}
Calibrating a polarimetric camera consists in determining the super-pixel matrix
$\mathbf{A}$ of \cref{eq:RealMueller} by solving \cref{eq:realPol-1} for all
the  super-pixels. To be able to solve this equation that has $\modified{4 \cdot 3=12}$ unknowns
in the matrix $\mathbf{A}$, at least three calibration light samples must be acquired
by the camera. Considering the general case where $N$ calibration samples are acquired
with $N\geq3$, the left hand side intensities vector of \cref{eq:realPol-1} becomes a
$4 \times N$ matrix, and the Stokes vector becomes a matrix of size $3 \times N$. The
matricial equation to solve is, thus, defined as:
\begin{equation}
\mathbf{I=AS,}
\label{eq:IntStokesRel}
\end{equation}
where $\mathbf{I}$ is the intensity matrix of the $N$ calibration light samples,
$\mathbf{A}$ is the super-pixel matrix, and $\mathbf{S}$ is the Stokes vectors matrix
of the $N$ calibration light samples. Consequently, using a least-squares approach,
the matrix $\mathbf{A}$ is equal to:
\begin{equation}
\mathbf{A=IS^{+},}
\label{eq:FinalSolution}
\end{equation}
where $\mathbf{S}^{+}$ is the pseudo-inverse of \textbf{S}. \modified{\cref{eq:FinalSolution}
constitutes the super-pixel calibration equation, and it can be solved for \textbf{A}
if the polarization states of the $N$ input calibration light samples are known}.
These states corresponds to the $N$ columns of the matrix:
\begin{figure}[!t]
\begin{centering}
\includegraphics[width=\linewidth]{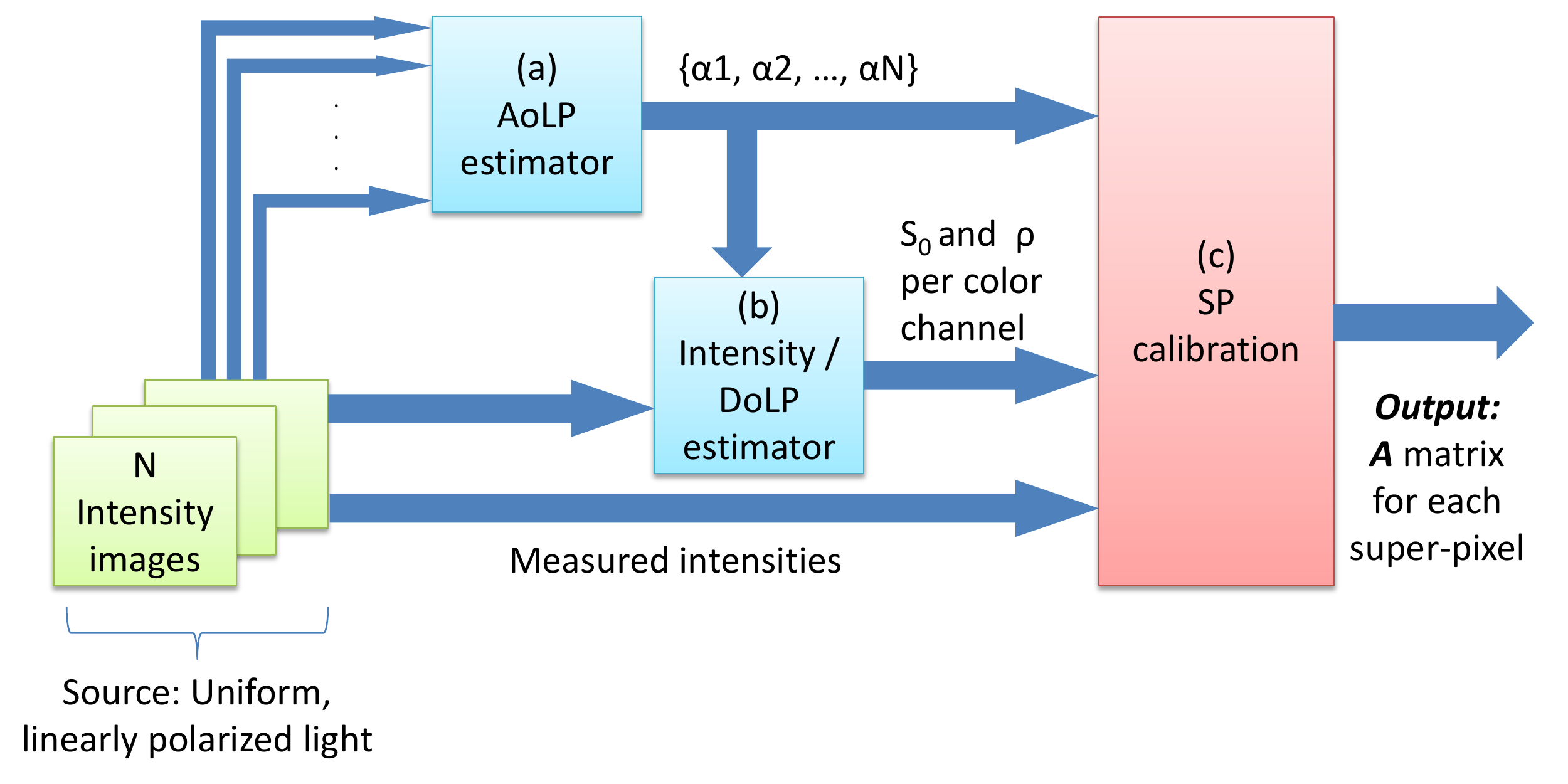}
\par\end{centering}\vspace*{-0.2cm}
\caption{\label{fig:CalibrationPipeline}\scriptsize Proposed calibration pipeline. From a uniform linearly
polarized light, $N$ samples are captured. The block (a) will estimate the angle
of linear polarization of each sample, and with them and the captured images, the
block (b) will estimate the degree of linear polarization and the intensity of the
source. Finally, with these estimations and the intensity measurements, the
super-pixel calibration is done (c).}\vspace*{-0.5cm}
\end{figure}
\begin{equation}
\mathbf{S}=\left[\begin{array}{ccc}
S_{01}                                    & ... & S_{0N} \\
S_{01} \rho_{1} \cos\left(2\alpha_1\right) & ... & S_{0N} \rho_{N} \cos\left(2\alpha_N\right) \\
S_{01} \rho_{1} \sin\left(2\alpha_1\right) & ... & S_{0N} \rho_{N} \sin\left(2\alpha_N\right) \\
\end{array}\right],
\label{eq:StokesInputSamples}
\end{equation}
where $S_{0n}$ is the intensity of the $n^{th}$ calibration light sample, $\rho_{n}$
is its degree of linear polarization (DoLP), and $\alpha_{n}$ is its angle of linear
polarization (AoLP), for $n=1,...,N$. These parameters
can be obtained with high accuracy but at the expense of a complex laboratory
set-up and time consuming experiments. In the following sections, we will show
how to estimate them using a uniform linearly
polarized (ULP) light that has moderate values in each of the three RGB channels. 

For creating this type of light, two configurations are possible, for which the
proposed calibration algorithm is valid: i) a linearly polarized light emitting
device is fixed, and the camera is rotated to obtain samples at different angles
of linear polarization, or ii) a rotative linear polarization filter is placed
between a fixed camera and a fixed unpolarized light source. Due to equipment
availability, the second configuration is used for the experiments in this paper.

\subsection{\label{subsec:AoLPEst}Input light angle of polarization
estimation}
In this section, we describe how to estimate the angles of linear polarization
of the $N$ calibration light samples represented by $\alpha_n$ with $n = 1, ..., N$
in \cref{eq:StokesInputSamples}. For the visible light wavelengths, the polarization
angle measured by a single super-pixel can be considered independent of the color.
Furthermore, if the camera is illuminated by a ULP light, all the
super-pixels should theoretically observe the same polarization angle. However, this
is not the case for a real camera, in general. Due to parameters dispersion, there is
not a single pixel in the sensor that can provide an accurate measurement of this
angle. Nonetheless, if the mean of the distribution of the AoLP errors is zero, then the mean of all the estimations should be
close to the true value. 

\modified{
To compute this mean AoLP, we select a certain number of pixels that are
negligibly affected by several undesired effects such as vignetting due to
the lens and the aperture, and polarization state errors due to light rays
of large angle of incidence. These pixels are those that are found in the central
region and that receive light rays pertaining to a small solid angle or Angular
Field Of View ($AFOV$). The relationship between the $AFOV$, the length $h$ in $mm$
of a square Region Of Interest (ROI) and the focal length $f$ of the lens in $mm$
is given by
\begin{equation}
AFOV = 2 \arctan\left(\frac{h}{2f}\right).
\label{eq:Afof}
\end{equation}
Let $p$ be the size of a super-pixel in $mm$ and $AFOV_{max}$ the maximum angular
field of view. \cref{eq:Afof} can be rearrange to obtain an upper limit for the
maximum size of the central region denoted by $N_{sp}$.
\begin{equation}
N_{sp} \leq \left\lfloor \dfrac{h}{p} \right\rfloor = \left\lfloor \dfrac{2 f}{p} \tan\left( \dfrac{AFOV_{max}}{2}\right) \right\rfloor
\label{eq:NbSuperPixels}
\end{equation}
Therefore, any small region of size $N_{sp}\times N_{sp}$ super-pixels around the
center that satisfies \cref{eq:NbSuperPixels} for an acceptable angular field of
view of about $1^\circ$ to $2^\circ$ can be considered  for estimating the AoLP
of the $n^{th}$ light sample using contiguous $2\times2$ pixels with the same color
filter.
} The measurements will comply with \cref{eq:realPol-1}, in which:

\begin{itemize}
\item $\modified{\mathbf{\underline{I}}=\left[I_{0}^{j},I_{45}^{j},I_{90}^{j},I_{135}^{j}\right]^{T}}$
is the  intensity vector of the $j^{th}$ super-pixel,
\item $\modified{\mathbf{\underline{S}}\simeq\mathbf{\hat{\underline{S_{n}}^{j}}}=\left[\hat{S_{n0}^{j}}, \hat{S_{n1}^{j}}, \hat{S_{n2}^{j}}\right]^{T}}$
\modified{is the Stokes vector of the incoming light sample measured by the $j^{th}$ super-pixel,}
\item $\left(T_{i}^{j},P_{i}^{j},\theta_{i}^{j}\right)$
are the parameters of a pixel that belongs to the super-pixel
$j$, and with a micro-polarizer oriented at an angle of $i$ degrees.
\end{itemize}

For the central pixels, negligible influence of the lens and small sensor artifacts are assumed. Then, the ideal values of
$T_{i}^{j}=0.5$, $P_{i}^{j}=1$ \modified{for all
$i=\left\{0^\circ,45^\circ,90^\circ,135^\circ\right\}$}
and $\theta=\left\{ 0{^\circ},45{^\circ},90{^\circ},135{^\circ}\right\}$ can be adopted
as a good approximation. Consequently, the matrix \textbf{A} is completely known,
and with its pseudo-inverse, the Stokes vector of the $n^{th}$ light sample, \modified{
$\mathbf{\hat{\underline{S_{n}}^{j}}}$ }can be obtained by:
\begin{equation}
\modified{
\mathbf{\hat{\underline{S_{n}}^{j}}}=\left[
\begin{array}{ccc}
\hat{S_{n0}^{j}} & \hat{S_{n1}^{j}} & \hat{S_{n2}^{j}} \\
\end{array}
\right]^{T}=\mathbf{A^{+}I}.}
\end{equation}
The AoLP $\alpha_{j}$ measured by the super-pixel $j$ is then given by:
\begin{equation}
\modified{
\alpha_{j}=\dfrac{1}{2}\arctan\left(\dfrac{\hat{S_{n2}^{j}}}{\hat{S_{n1}^{j}}}\right).
\label{eq:AoPEstimation}
}
\end{equation}
\modified{This operation is repeated for all the $K=N_{sp}\cdot N_{sp}$
super-pixels in the central region. Due to the periodicity of the AoLP, the
average of the $K$ angles might conduct to wrong results. Thus, their circular average
is computed instead according to \cref{eq:circular_avg_formula}.}
This yields the angle $\hat{\alpha_{n}}$, which is an estimation of the AoLP of the
$n^{th}$ light sample denoted by $\alpha_n$ with $n = 1, ..., N$ in \cref{eq:StokesInputSamples}.

\modified{
\begin{equation}
\begin{array}{c}
\sin\left(2\hat{\alpha_{n}}\right)=\frac{1}{K} \sum_{j=1}^{K} \sin\left(2\alpha_{j}\right) \\
\cos\left(2\hat{\alpha_{n}}\right)=\frac{1}{K}\sum_{j=1}^{K} \cos\left(2\alpha_{j}\right) \\
\hat{\alpha_{n}}=\frac{1}{2}\arctan\left(\sin\left(2\hat{\alpha_{n}}\right) / \cos\left(2\hat{\alpha_{n}}\right)\right)
\end{array}
\label{eq:circular_avg_formula}
\end{equation}
}

\subsection{\label{sec:IntensityEst}Light samples intensity and DoLP estimation}

We will now describe how to estimate the
light intensities and the DoLP of the $N$ calibration light samples defined
in the matrix $\mathbf{S}$ of \cref{eq:StokesInputSamples}.

For the same reasons as explained in \cref{subsec:AoLPEst}, a \modified{$N_{sp}\times N_{sp}$ } 
super-pixels region around the center of the sensor is  considered for all the light samples that will have the same intensity and the same DoLP. Only the AoLP will be different by selecting a different orientation of the rotative filter for each sample. Thus, in \cref{eq:StokesInputSamples}, $S_{01}=...=S_{0N}=S_{0}$ and
$\rho_{1}=...=\rho_{N}=\rho$. Consequently, the Stokes matrix of the $N$ calibration light
samples \textbf{S} can be split into two matrices: a $3 \times 3$ matrix \textbf{L} that only
depends on $\left(S_{0},\rho\right)$, and a $3 \times N$ matrix \textbf{G} that only
depends on the angles of linear polarization $\alpha_{n}$ estimated in  \cref{subsec:AoLPEst},
such that $\mathbf{S=LG}$:
\begin{equation}
\begin{array}{c}
\mathbf{S}=
    \left[\begin{array}{ccc}
        S_{0} & 0 & 0 \\
        0 & S_{0} \rho & 0 \\
        0 & 0 & S_{0} \rho \\
    \end{array}\right]
    \left[\begin{array}{ccc}
        1                         & ... & 1 \\
        \cos\left(2\alpha_1\right) & ... & \cos\left(2\alpha_N\right) \\
        \sin\left(2\alpha_1\right) & ... & \sin\left(2\alpha_N\right) \\
    \end{array}\right]\\
    \\
\end{array}
\label{eq:splitStokes}
\end{equation}

Combining \cref{eq:IntStokesRel}, which is the super-pixel calibration equation, and
\cref{eq:splitStokes}, yields:
\begin{equation}
\mathbf{IG^+=AL,}
\label{eq:MatrixMWithStokes}
\end{equation}
where \textbf{I} is the $4 \times N$ matrix of the measured intensities.
For the super-pixel $j$, each row $i$ of the result $\mathbf{IG^+}$ can be
expressed as:
\begin{equation}
\left(\mathbf{IG^+}\right)_{i}^{j}=
\left[\begin{array}{ccc}
X_{i}^{j} & Y_{i}^{j} & Z_{i}^{j}
\end{array}
\right],
\label{eq:SingleLineParam}
\end{equation}
where 
$X_{i}^{j}=\frac{T_{i}^{j}S_{0_{i}}^{j}}{P_{i}^{j}}$,
$Y_{i}^{j}=T_{i}^{j}S_{0_{i}}^{j} \rho_{i}^{j} \cos\left(2\theta_{i}^{j}\right)$,
and $Z_{i}^{j}=T_{i}^{j}S_{0_{i}}^{j} \rho_{i}^{j} \sin\left(2\theta_{i}^{j}\right)$.
If the camera is considered ideal for the central pixels, as in the
previous section, each of the four rows of $\mathbf{IG^+}$ allows to calculate a pair
$\left(S_{0_{i}}^{j},\rho_{i}^{j}\right)$ as follows:
\begin{equation}
\begin{array}{cc}
S_{0_{i}}^{j}=2X_{i}^{j} & \rho_{i}^{j}=\dfrac{\sqrt{Y_{i}^{j2}+Z_{i}^{j2}}}{X_{i}^{j}}.
\end{array}
\label{eq:S0Calc}
\end{equation}

Repeating this procedure for the $K$ super-pixels and the $N$ samples will
yield two sets of \modified{$R = N_{sp}\cdot N_{sp}\cdot 4$ }
intensities and DoLP: $\left\{ S_{0}^{1},S_{0}^{2},...,S_{0}^{R}\right\} $ and
$\left\{ \rho^{1},\rho^{2},...,\rho^{R}\right\}$. From these two sets, the light
parameters, $S_0$ and $\rho$, can be estimated by extracting either the maximum
(highly sensitive to noise and outliers), the average (affected by lens vignetting
and outliers) or the median (affected only by lens vignetting) value. Because of its
robustness to outliers, the median value has been chosen and implemented for the experiments.

It is important to note that a color camera is used. To be free from
the requirement of using a white light, the detected intensities and DoLP
have to be classified per color channel, without mixing them. The color
channel to which a super-pixel belongs to is given by its position $j$.

At this point, an estimation of the light intensity $\hat{S_{0}}$, the degree of
linear polarization $\hat{\rho}$ per color channel, and the angle of polarization
$\hat{\alpha_{n}}$ at the $n^{th}$ position of the linear filter has been obtained.
Therefore, the Stokes matrix can be built, as in \cref{eq:EstimatedStokes},
\begin{equation}
\hat{\mathbf{S}}=\left[
\begin{array}{ccc}
\hat{S_{0}} & 0 & 0 \\
0 & \hat{S_{0}} \hat{\rho} & 0 \\
0 & 0 & \hat{S_{0}} \hat{\rho} \\
\end{array}
\right]
\left[
\begin{array}{ccc}
1 & ... & 1 \\
\cos\left(\hat{\alpha_{1}}\right) & ... & \cos\left(\hat{\alpha_{N}}\right) \\
\sin\left(\hat{\alpha_{1}}\right) & ... & \sin\left(\hat{\alpha_{N}}\right) \\
\end{array}
\right].
\label{eq:EstimatedStokes}
\end{equation}

\noindent and its pseudo-inverse $\hat{\mathbf{S}}^{+}$ computed and used in
\cref{eq:FinalSolution} to calculate the $j^{th}$ super-pixel matrix that we
will denote here by $\hat{\mathbf{A_{j}}}$. It follows that, from
$\hat{\mathbf{A_{j}}}$, each row allows to compute the parameters
$\left(T_{i},P_{i},\theta_{i}\right)$ for each of the four pixels that
compose the $j^{th}$ super-pixel.

\section{\label{sec:Experimental-results}Experiments}
\modified{
Our experimental setup is composed of a Basler acA2440-75ucPOL camera with a Sony Polarsens
IMX250MYR sensor of pixel size equal to $3.45 \mu m$ or super-pixel size equal to
$6.9\mu m$, and a Fuji-film HF16XA-5M - F1.6/16mm lens. To compute an initial estimation
of the AoLP and DoLP required by our calibration method, we have chosen a central region
of  $50\times 50$ super-pixels determined according to \cref{eq:NbSuperPixels}. This region
corresponds to incident light rays with a maximum angle of incidence of $0.625^\circ$ that
is relatively small and will give a good initial estimation of the AoLP and DoLP.

}

\begin{figure}
\begin{centering}
\includegraphics[width=8.5cm]{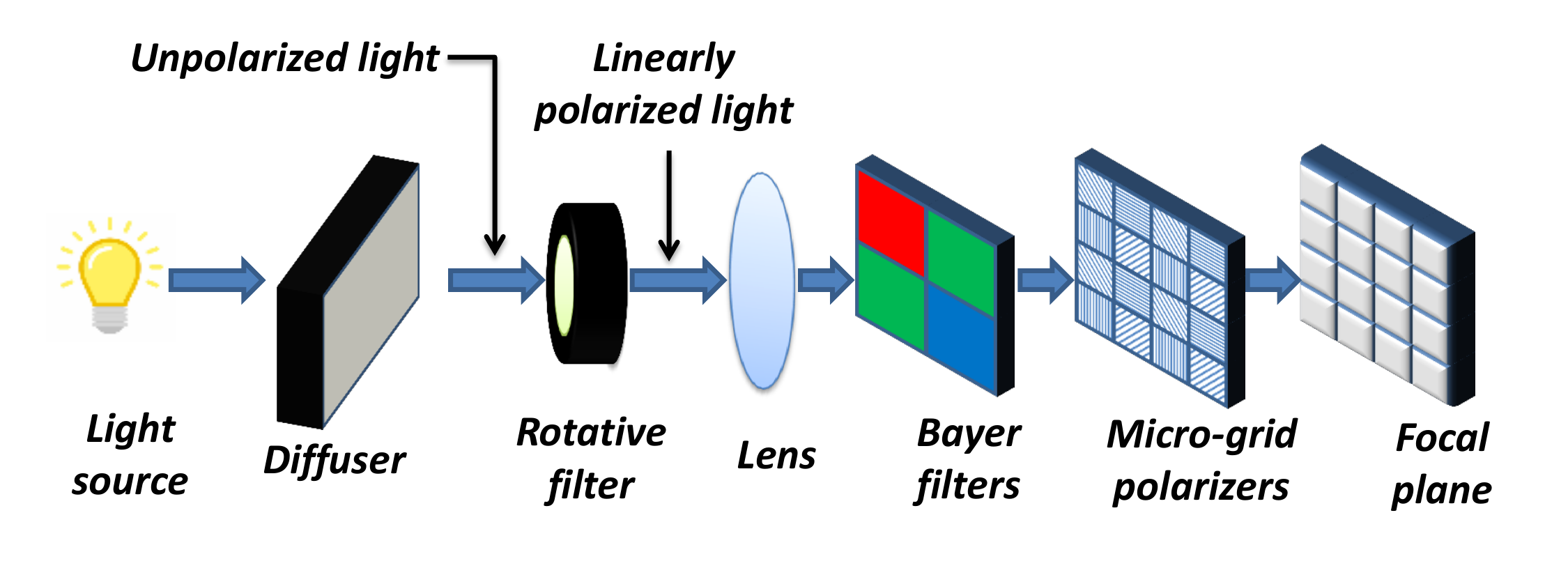}
\par\end{centering}\vspace*{-0.2cm}
\caption{\label{fig:Experiment-model-and}\scriptsize The experiment set up where only one super-pixel is represented.}
\vspace*{-0.4cm}
\end{figure}
The developed algorithm runs on a computer with Intel Core
i7-10850H @ 2.7 GHz and 32 GB of RAM. The OS is Ubuntu
18.04 LTS 64 bits. The program runs in 7 seconds for 7 samples, and
8 seconds for 73 samples approximately. The experimental set-up model is
shown in \cref{fig:Experiment-model-and}. The uniform, unpolarized light source
device is a Schott Fostec DCR III fiber optic illuminator, with a Schott ColdVision
back light A08927. A 50mm linear glass polarizing filter is used (Edmund Optics
Inc \#56-329), mounted on a metric polarizer mount (Edmund Optics Inc \#43-787).
The linear polarizer filter is rotated by hand. Each position
of the filter corresponds to a light sample, and for each sample ten acquisitions of
the light are done and averaged to reduce the effects of the noise in the parameters
estimation. The acquisitions are done in a dark room to reduce the influence of the
environment. Furthermore, the recommendations given in \cite{25_Connor} have been
followed. \modified{Particularly, the lens has been correctly focused at the  light
source, and the $f$-number has been set higher than 2.8 for all the experiments.
Finally, we have acquired several images with the camera in total darkness and verified that, with
a 12-bit pixel count and an exposure time of 1s, the dark current can effectively be neglected as
reported in \cite{25_Connor}.}
\begin{figure}
\begin{centering}
\includegraphics[width=7.5cm]{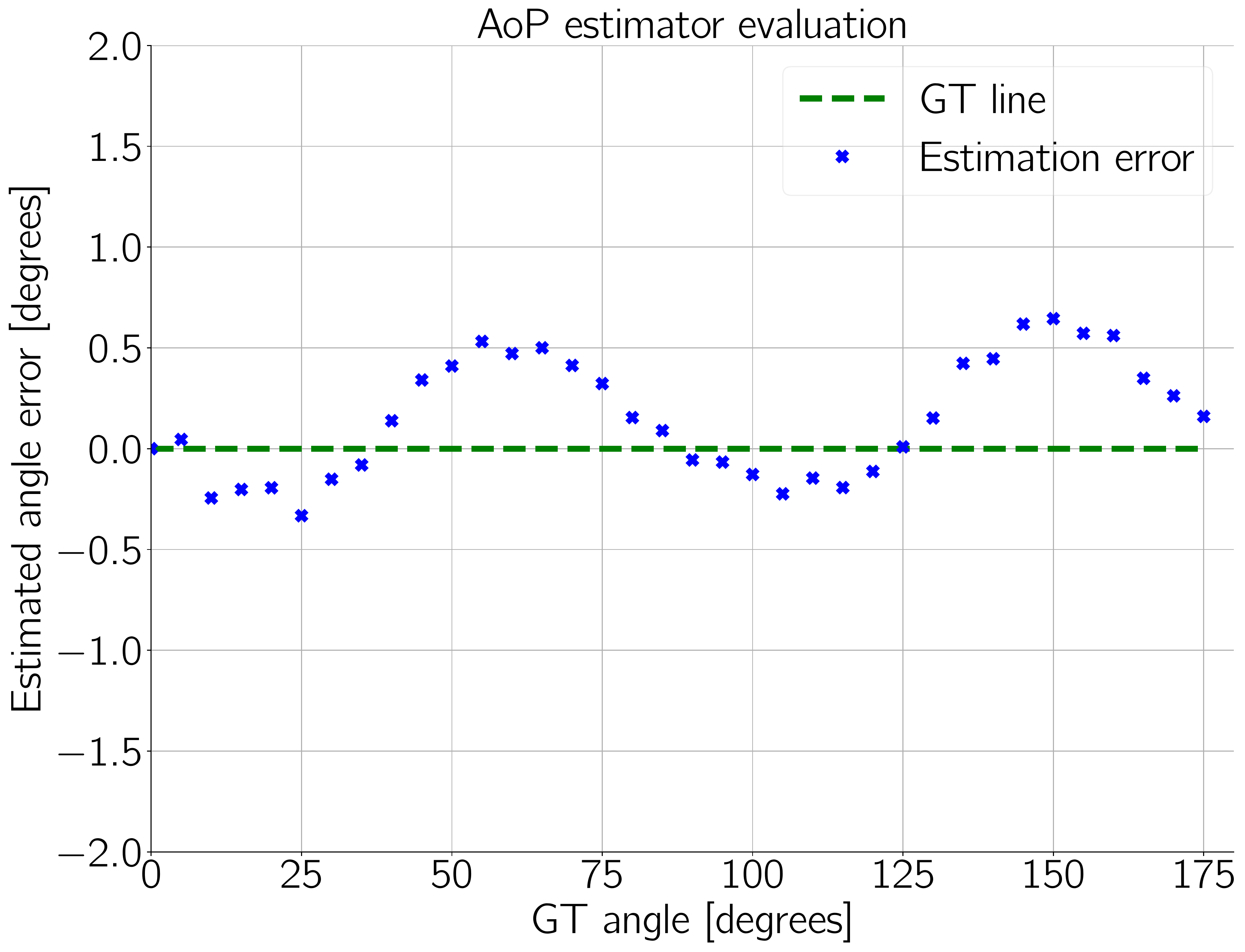}
\par\end{centering}\vspace*{-0.2cm}
\caption{\label{fig:Angle-of-polarization}\scriptsize AoLP estimator evaluation.
The maximum error is of $0.65^\circ$, and the RMSE is of $0.3316^\circ$
for all the range from $\left[0^\circ,180^\circ\right]$. The sine-like evolution
is mainly due to the error in the polarizers orientations \cite{GitHubRepo}.}\vspace*{-0.5cm}

\end{figure}

The first experiment that we have done is to verify the quality of the estimated AoLP
with the camera. To do so, different light samples with different AoLP have
been used. The AoLP values are set by turning the polarization filter in front of the
light source by steps of $5^\circ$, in the range $\left[0^\circ,175^\circ\right]$. 
\modified{Thirty-six images of the source light samples are acquired with the camera,
and from them the AoLP are calculated and compared to the true reference values given by
the position of the polarization filter. For better visualization, only the deviations from
the reference values are represented in \cref{fig:Angle-of-polarization}.
In case the estimated AoLP values are the same as the reference values, an horizontal line
at zero degree is obtained. However, in our case, the AoLP error curve exhibits a sine-like
shape that is due to errors in the parameters of the camera. Indeed, it can be proven that a
small error in the parameters of the camera due to imperfections will induce, in first order
approximations, four error terms in the expression of the estimated AoLP. These error terms are
functions of the sine and cosine of the true AoLP and they appear in the expressions of the
computed Stokes components $S_1$ and $S_2$. Because of these additional components, and that
the ratio of these two Stokes components is proportional to the tangent of the AoLP
(\cref{eq:PolarizationParams}), the error curve follows a sine and cosine rule.
The detailed demonstration of this effect can be found in the supplementary material \cite{GitHubRepo}
(it is not included in the main paper since it is an auxiliary result of this work, and also
due to space constraints).} Nonetheless, by considering pixels around the center,
and averaging several samples, the estimation error is reduced, such that the RMSE is $0.3316^\circ$,
and the maximum error is $\pm 0.65^\circ$ in all the range. Hence, the experiment confirms that the
camera can be used to provide reliable measurements of the AoLP of the ULP light. Additionally, it avoids
the requirement of aligning the rotative filter and the camera, since the measurements are already
in the camera's coordinate frame.

The next step is to evaluate the accuracy of the calculated intensities, AoLP and DoLP
with the uncalibrated and the calibrated camera with $N$ number of calibration light samples.
Prior to the test, a database has been created with all the images of the required calibration
light samples. These samples have the same intensity and DoLP, but different
AoLP in the range of $\left[0^\circ,180^\circ\right]$. For the test, $N$ sample
images are randomly selected from the database and used to compute the pixel parameters.
Once the camera is calibrated, the intensity, DoLP and AoLP of a test image are calculated.
The mean and standard deviation of the test light parameters are calculated over all the
sensor area. This test is repeated several times for each value of $N=3, ..., 36$.
For each run of the algorithm, a new set of $N$ random calibration images is chosen to
calibrate the camera. The GT values of the test image are:  $S_{0}=1437$, $\rho=0.97$, and
$\alpha=43{^\circ}$. 
For space reasons, only the results for the red channel are shown in \cref{fig:Comparative-graphs}. Those for the other
channels can be found in the supplementary  material \cite{GitHubRepo}.

\begin{figure*}[t]
\begin{centering}
\begin{tabular}{ccc}
\includegraphics[width=5cm]{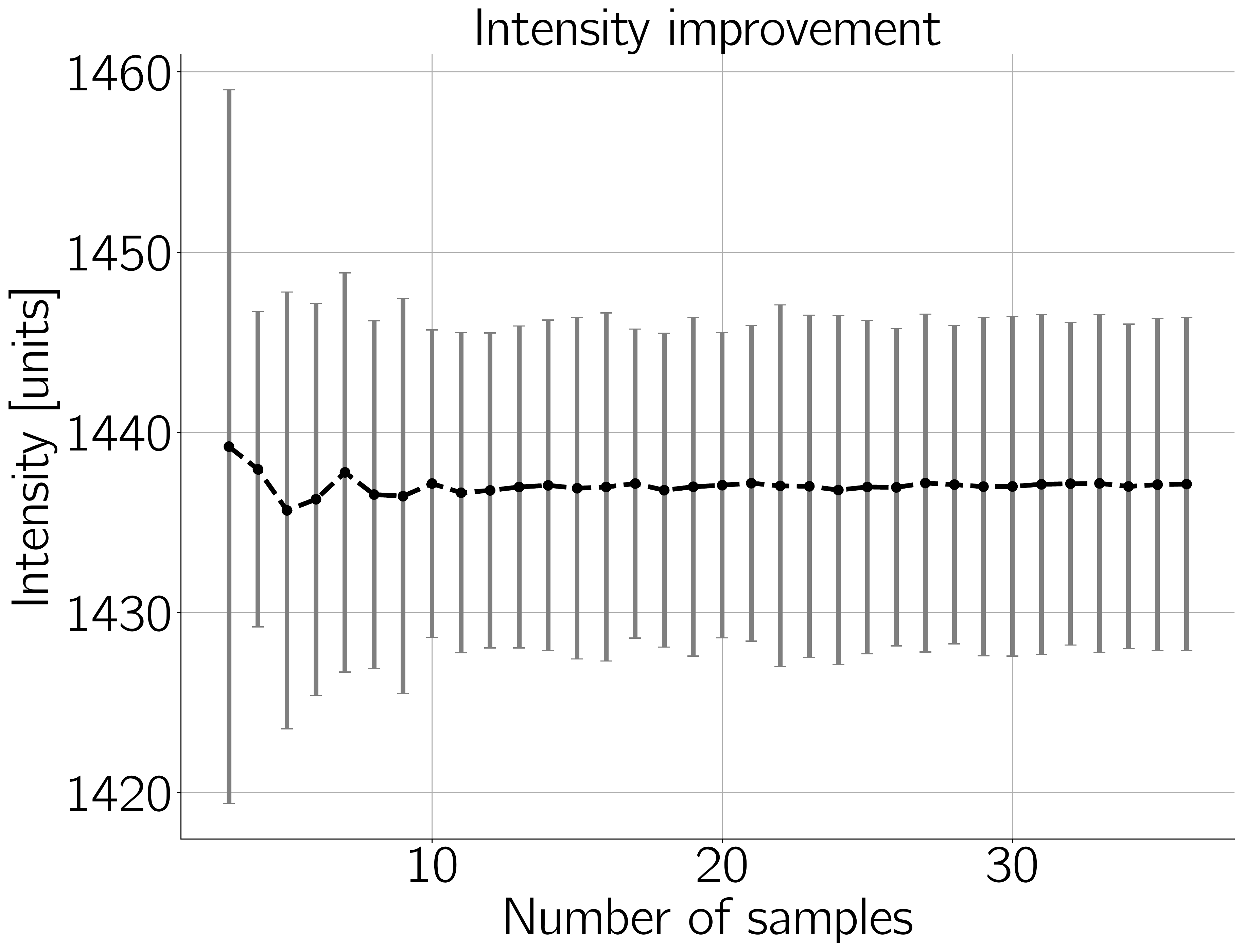} & \includegraphics[width=5cm]{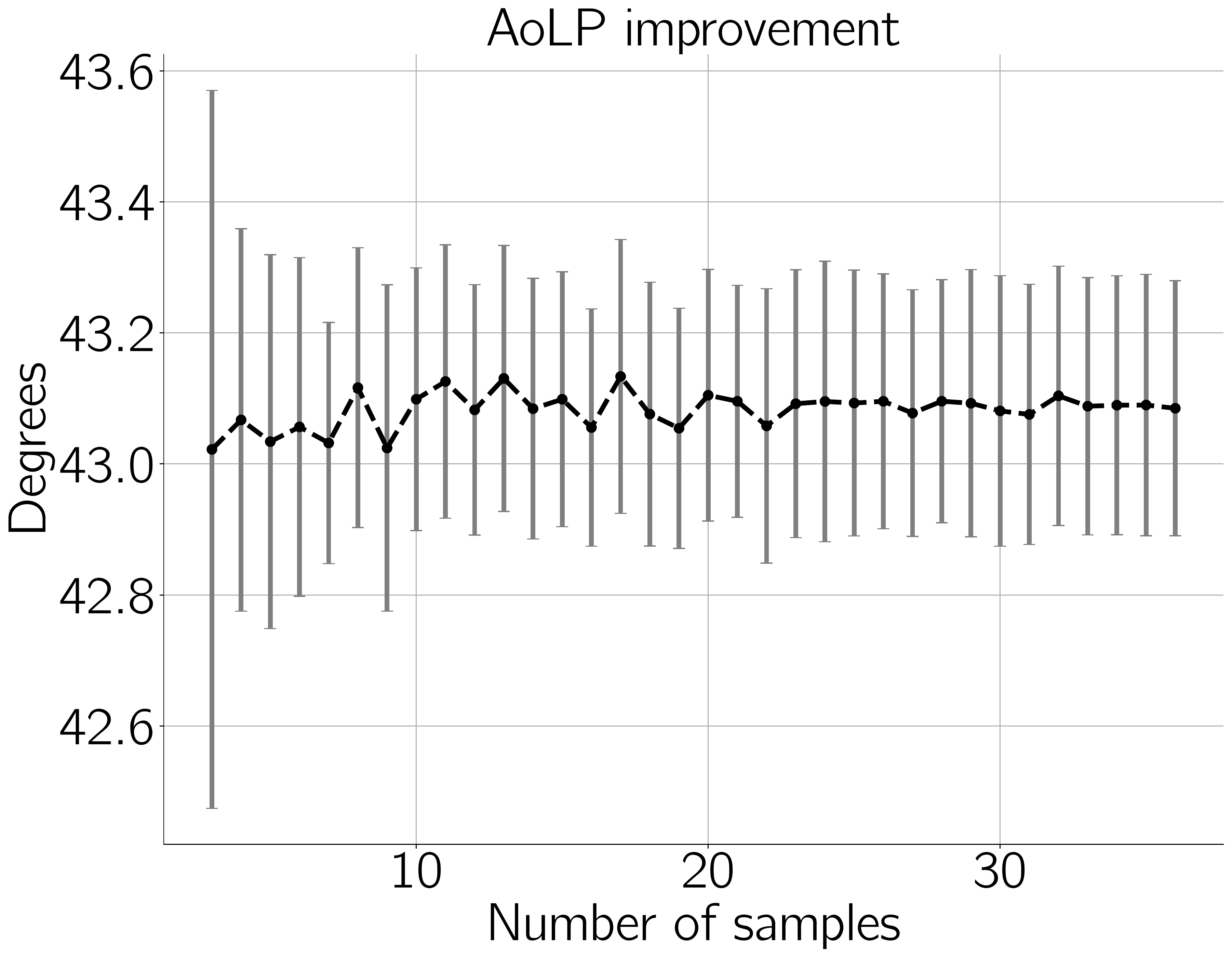} & \includegraphics[width=5cm]{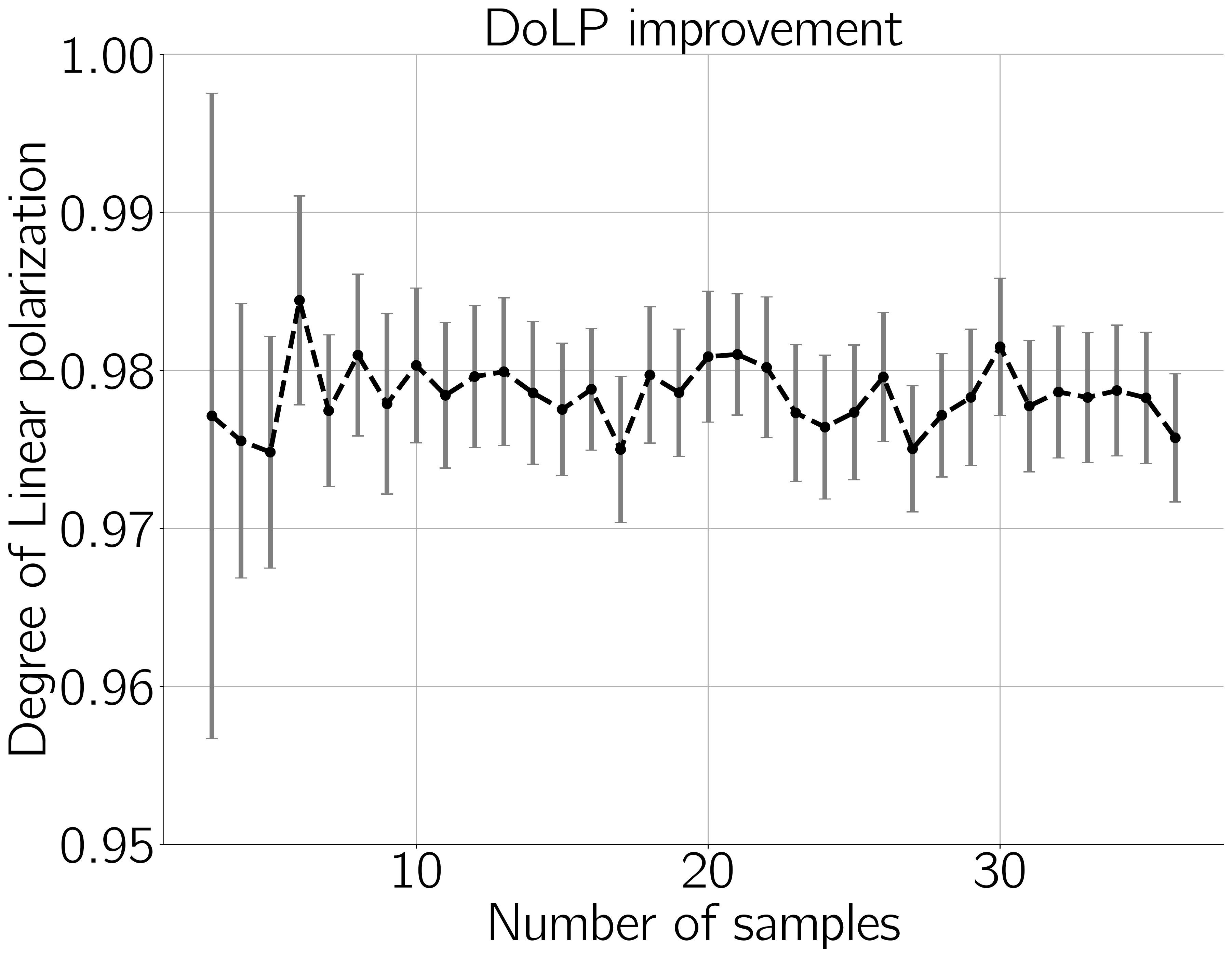} \tabularnewline
(a) & (b) & (c)\tabularnewline
\end{tabular}
\par\end{centering}\vspace*{-0.2cm}
\caption{\label{fig:Comparative-graphs}\scriptsize Comparative graphs of how the accuracy
changes with the number of light samples. Each sample is taken at a different
position of the rotative filter. Each point in the curve is the average of several
runs of the algorithm. The black curves are the mean values, and the gray vertical
bars are the standard deviation. When uncalibrated, the parameters are:
$S_{0}=1336.327\pm61.731$, $\rho=0,9776\pm 0,006$, and $\alpha=43,024^\circ\pm 0.48^\circ$.
When calibrated and the amount of samples $N\geq10$, $S_{0}=1437.134\pm 9.25$,
$\rho=0.98\pm0.005$ and $\alpha=43.099^\circ\pm0.2^\circ$.
(a) Intensity $S_{0}$. (b) AoLP $\alpha$. (c) DoLP $\rho$.}

\end{figure*}

As shown in \cref{fig:Comparative-graphs}, when five or more calibration light samples are
used, the standard deviation is considerably reduced with respect to the case
when only three samples are used, and when $N\geq10$, the values are stabilized.
More precisely, for $N\geq10$, $S_{0}=1437.134\pm 9.25$, $\rho=0.98\pm0.005$
and $\alpha=43.099^\circ\pm0.2^\circ$. The same test image has been used
with the uncalibrated camera , and the obtained parameters were:
$S_{0}=1336.327\pm61.731$, $\rho=0,9776\pm 0,006$, and
$\alpha=43,024^\circ\pm 0.48^\circ$. This experiment also corroborates
that the camera calibrated with our algorithm reduces the disparity between
values over the sensor area with respect to the uncalibrated
camera.

To confirm the validity of our method with respect to other algorithms, the
calibration results have been compared with the super-pixel (SP) method
described in \cite{22_Yilbert_2020, 18_Powell_Gruev}. \cref{tab:Yilbert_comparison}
summarizes the comparison results. The AoLP used for the SP method has been
measured from the rotative filter, while for our method they have been estimated.
The difference between the mean values of the intensity and the DoLP is expected,
since each method uses a different reference during calibration. However, the most
important results are the standard deviations that reflect how similar the
measurements of the ULP light are after the correction over
the entire sensor area. One can notice that the results obtained by both approaches have
similar accuracy. However, our method has the advantage of being experimentally simple:
it does not require any specific devices to measure the light polarization state. Additionally, the
time required to take the samples is reduced for the user since he only needs to randomly
turn the polarizer a few times. Also, the measurements of the orientation from the rotative
filter are not required since the algorithm will estimate them automatically.

\begin{table}
\begin{centering}
\begin{tabular}{|p{1.5cm}|c|c|c|}
\hline
& {\centering{}{\footnotesize{}$S_{0}$}} & {\centering{}{\footnotesize{}$AoLP$}} & {\centering{}{\footnotesize{}$DoLP$}}\tabularnewline
\hline
{\centering{}{\footnotesize{}Uncalibrated}} & {\centering{}{\footnotesize{}3399.2 [65.475]}} & {\centering{}{\footnotesize{}59.983 [0.215]}} & {\centering{}{\footnotesize{}0.9863 [0.0041]}}\tabularnewline
\hline
{\centering{}{\footnotesize{}SP method}} & {\centering{}{\footnotesize{}3402.4 [10.723]}} & {\centering{}{\footnotesize{}60.035 [0.128]}} & {\centering{}{\footnotesize{}1.005 [0.004]}}\tabularnewline
\hline
{\centering{}{\footnotesize{}Our method}} & {\centering{}{\footnotesize{}3298.0 [10.402]}} & {\centering{}{\footnotesize{}59.701 [0.128]}} & {\centering{}{\footnotesize{}0.985 [0.004]}}\tabularnewline
\hline

\end{tabular}\vspace*{0.2cm}
\caption{\label{tab:Yilbert_comparison}\scriptsize Comparison of our method
with the super-pixel (SP) method \cite{22_Yilbert_2020}. Content of each
cell: mean value [standard deviation]}\vspace*{-1cm}
\end{centering}
\end{table}

\begin{figure*}
\begin{centering}
\begin{tabular}{cccc}
\raisebox{-.5\height}{\includegraphics[width=3.7cm]{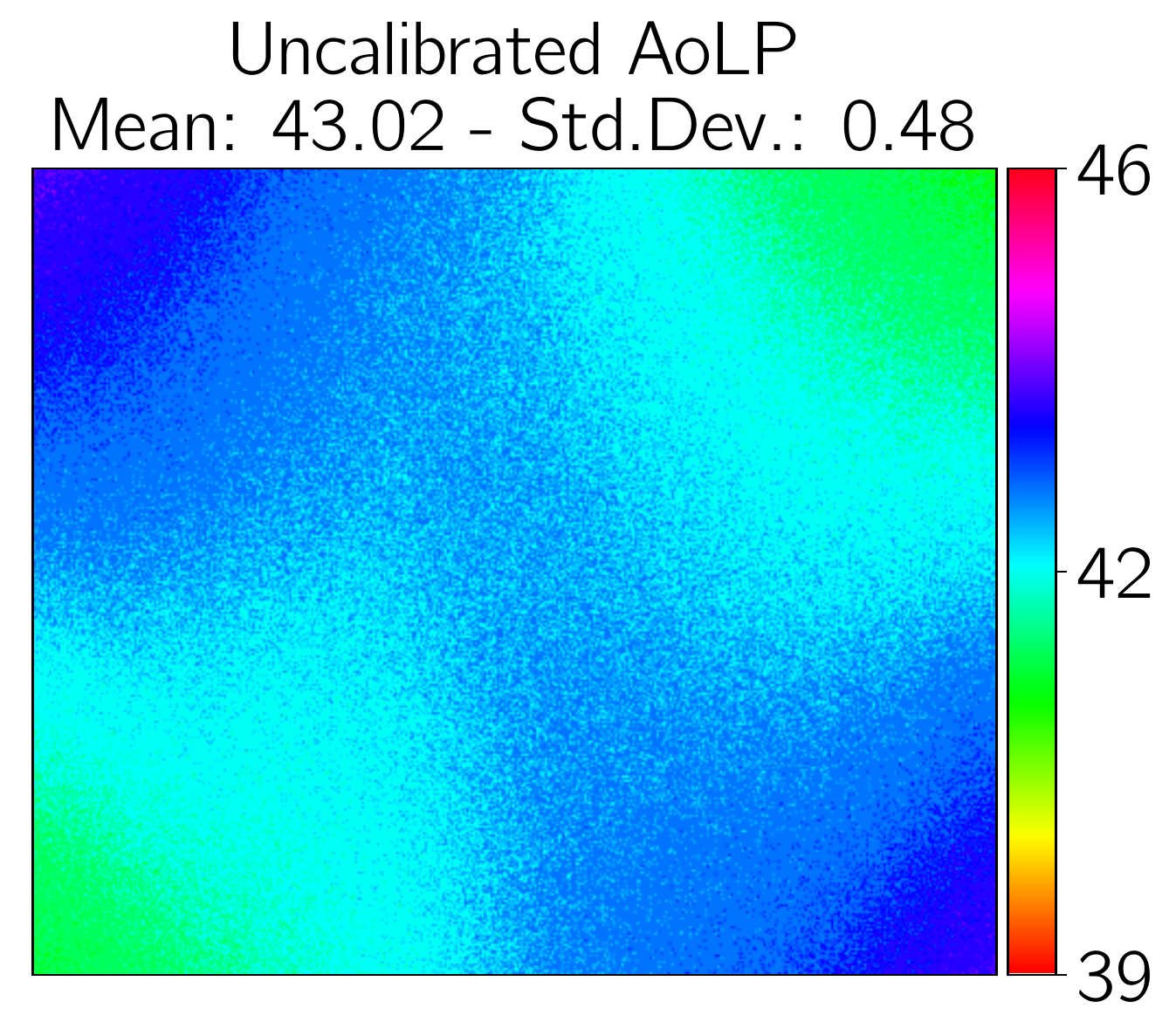}} & \raisebox{-.5\height}{\includegraphics[width=3.7cm]{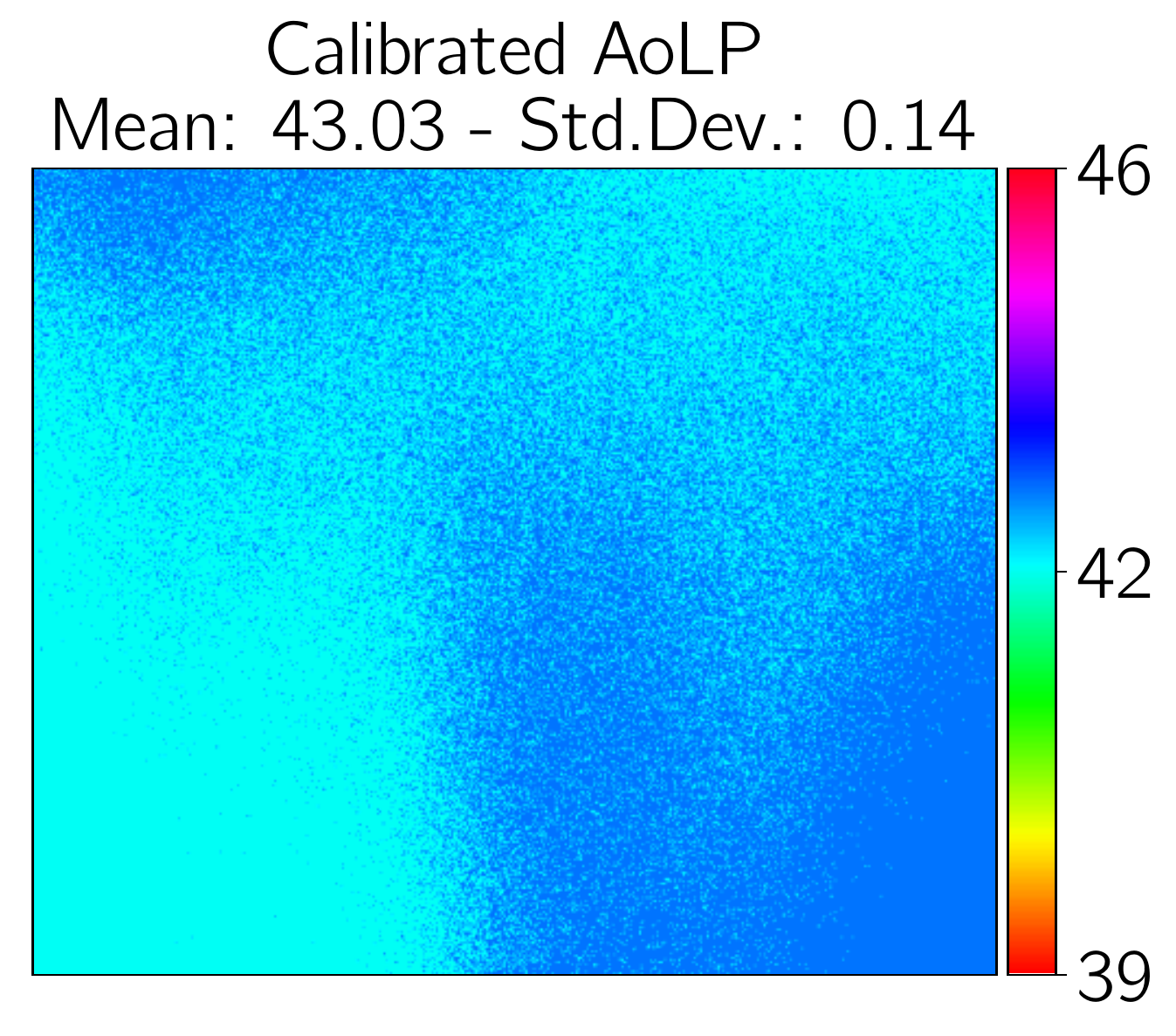}}  &
\raisebox{-.5\height}{\includegraphics[width=3.7cm]{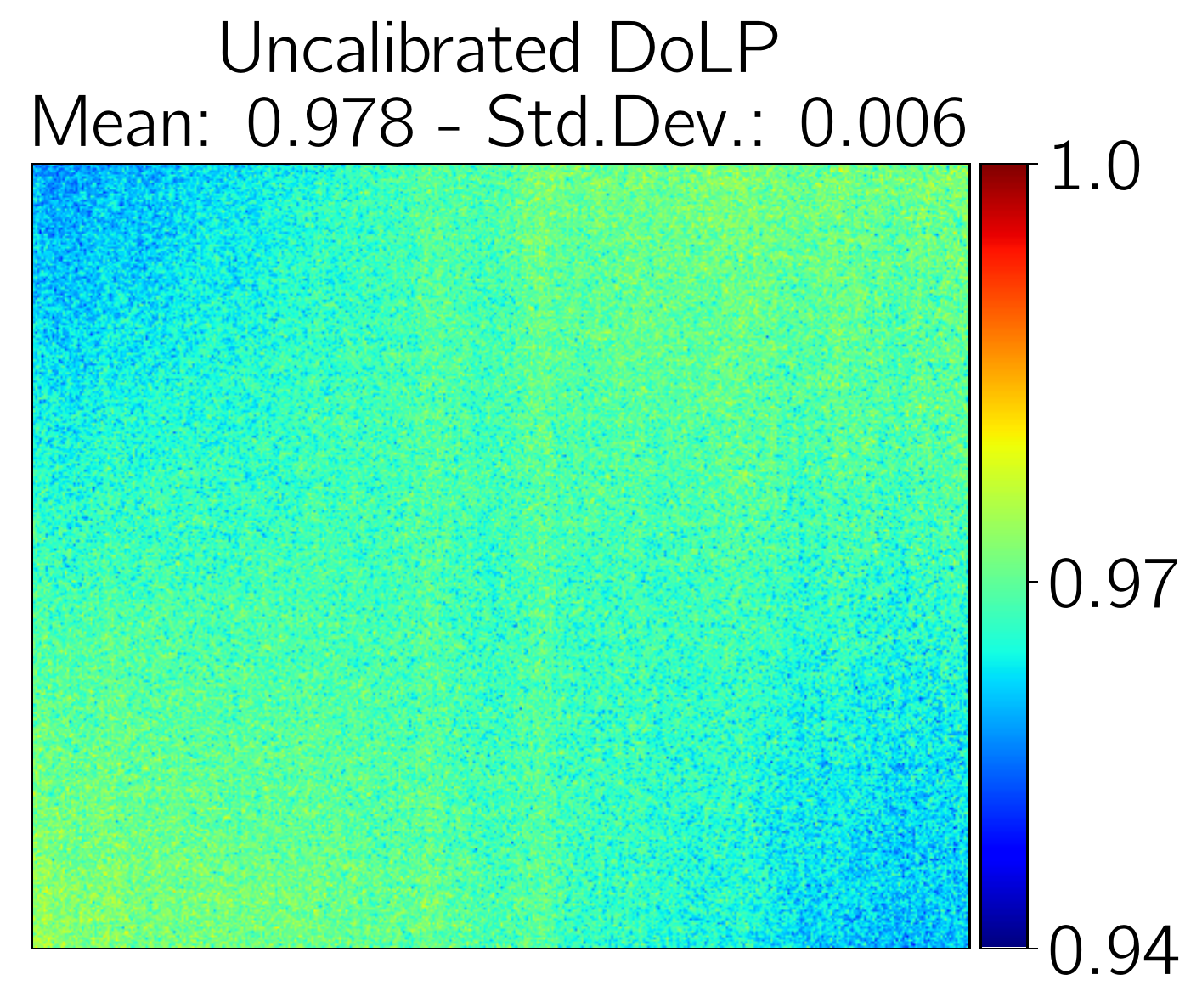}} & \raisebox{-.5\height}{\includegraphics[width=3.7cm]{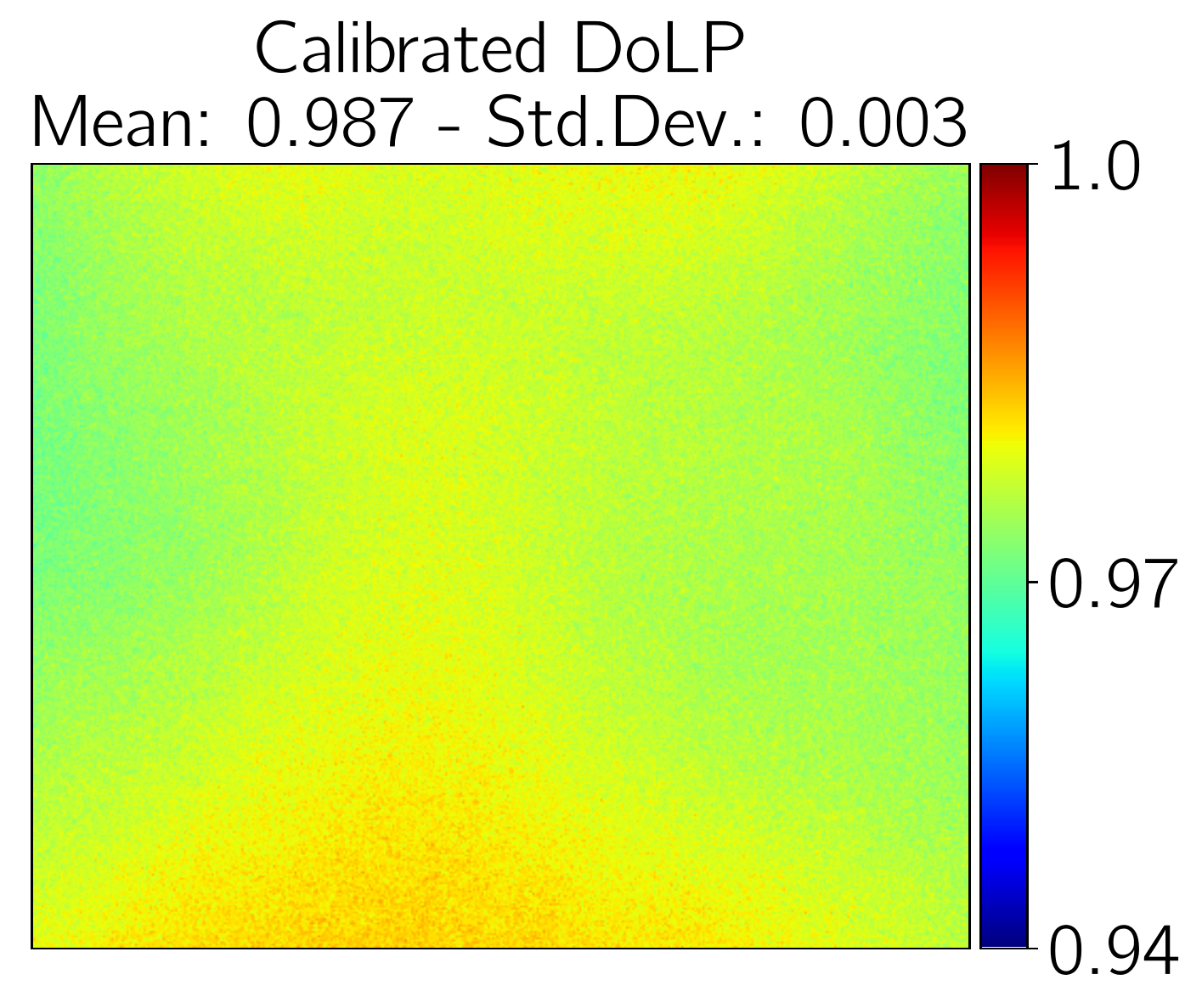}}\tabularnewline
(a) & (b) & (c) & (d) \tabularnewline
\end{tabular}
\par\end{centering}\vspace*{-0.2cm}
\caption{\label{fig:Comparision-of-the}\scriptsize\modified{Calibration improvement image for the red
channel. (a) Uncalibrated AoLP image (b) Calibrated AoLP image. (c) Uncalibrated DoLP
image. (d) Calibrated DoLP. AoLP uses the HSV color palette, and the DoLP uses the
Jet palette. Since the light source is ULP, the ideal response would be images
with a single value for all the pixels for the AoLP and another value for the DoLP.
Due to characteristics dispersion in the pixels and the presence of the lens,
the uncalibrated images present different values between the center of the sensor
and its borders in the two polarization parameters. After calibration,
the distribution of measured values is reduced. Even though there seems to be a difference
in the colors of the calibrated images, looking at the color bars range, it is clear
that the range of values present in the image is small compared to the uncalibrated cases.
Additionally, the mean and the standard deviation of values
in the images are detailed in the images titles.}}\vspace*{-0.4cm}
\end{figure*}

Finally, the effect of calibration on an image of a ULP light is presented in
\cref{fig:Comparision-of-the}. For space reasons, only the results for
the red channel are shown. The results for the other channels are available in
\cite{GitHubRepo}. In this figure, the images (a) and (b) corresponds to AoLP image,
before and after calibration, and the images (c) and (d) are the corresponding
images of the DoLP. These images are the measurements of the camera when it is illuminated
by a ULP light. For the uncalibrated case, the
values of the AoLP and the DoLP are in the intervals $\left[0.9465,1.0\right]$,
and $\left[41.099^\circ,44.879^\circ\right]$, respectively. For the calibrated
camera the corresponding values are in the intervals $\left[0.9741,1.0\right]$, and
$\left[41.934^\circ,43.824^\circ\right]$, respectively. What is important to note
in these images is not the color but the variations in the colors. Since, these
images represent the values of the measured AoLP and DoLP, by the
individual pixels of the camera, they should respectively be the same, i.e. all
the pixels of the same image should have the same color, due to the
fact that the observed light is uniform and linearly polarized. However, this is
not the case. The variations in the color represent the measurement
errors by the individual pixels. One can note that the color is more uniform (and
therefore, there is less errors) for the calibrated camera than for the uncalibrated camera.

Additionally, these images show that, for the uncalibrated images,
both polarization parameters present changes in the borders with respect to the
center of the image. This is a consequence of the lens vignetting. Indeed,
the variation of the polarization state of the incoming light depends on the point of
incidence of the light at the lens surface. As one can see in \cref{fig:Comparision-of-the},
the uncalibrated measurements have strong variations in the four corners of the
images, and in the center they are mostly constant. This effect is equivalent to having a Mueller
matrix applied to the input light, where this matrix is different at each lens point.
The proposed algorithm accounts for this alteration of the polarization state in the pixel
model, by modulating the polarizers parameters $P_{i}$ and $\theta_{i}$, to fit
the measurements with a flat-field response. $T_{i}$ has no effect over these
parameters since it is a factor of all the three Stokes parameters, thus it is
cancelled when computing the DoLP and the AoLP. However, $T_{i}$
corrects the vignetting effect over the intensity image, which has not been included
here due to space restrictions, but it can be found in the supplementary material
\cite{GitHubRepo}.

\modified{\noindent\textbf{Ablation study.} To test the influence of each module,
several scenarios have been evaluated and summarized in Tab. \ref{tab:ablationTable}.
The first row of this table, SC1, corresponds to the results when using the
uncalibrated camera. Then, in SC2, the AoLP estimator has been used and the DoLP
and intensity of the input light have been fixed to 0.8 and 2000, respectively.
As shown in the table, this modification changes the corresponding mean value measured
in the entire image, but their standard deviation (SD) is not modified. The advantage of having a module
that estimates these parameters is that an early saturation of the measured value is
avoided. Then, in SC3, the AoLP estimator is disabled, and these values have been
measured from the filter ruler. In this case, the AoLP after calibration presents a
slightly smaller SD than when using the estimator, due to the small error in the
measurements of this parameter. Nonetheless, this has as trade off a large experiment time, and the
requirement of a rotative mount with a ruler. In SC4, a similar experiment to SC3
is done, but a fixed shift is introduced in the AoLP measurements. From
Tab. \ref{tab:ablationTable} it can be seen that this affects the mean value of the
measured AoLP, but the SD remains low. This is normal, since the AoLP is relative
to the measurement system. Finally, our entire pipeline is tested in SC5, in which
a small SD is obtained in all the variables, and additionally, the mean values
are close to the GT values. Therefore, using a simple calibration set-up as ours can provide not only accuracy, but also precision in the measurements given by the
camera.

\begin{table}
\begin{centering}
\begin{tabular}{|p{1.5cm}|c|c|c|}
\hline
& {\centering{}{\footnotesize{}$S_{0}$ $\left[0, 4095\right]$}} & {\centering{}{\footnotesize{}$AoLP$ ($^\circ$)}} & {\centering{}{\footnotesize{}$DoLP$ $\left[0, 1\right]$}}\tabularnewline
\hline
{\centering{}{\footnotesize{}SC1}} & {\centering{}{\footnotesize{}2750.66 [153.46]}} & {\centering{}{\footnotesize{}60.08 [0.416]}} & {\centering{}{\footnotesize{}0.986 [0.0033]}}\tabularnewline
\hline

{\centering{}{\footnotesize{}SC2}} & {\centering{}{\footnotesize{}1985.63 [4.88]}} & {\centering{}{\footnotesize{}59.87 [0.126]}} & {\centering{}{\footnotesize{}0.797 [0.00362]}}\tabularnewline
\hline

{\centering{}{\footnotesize{}SC3}} & {\centering{}{\footnotesize{}2967.34 [7.24]}} & {\centering{}{\footnotesize{}59.61 [0.125]}} & {\centering{}{\footnotesize{}0.987 [0.0045]}}\tabularnewline
\hline

{\centering{}{\footnotesize{}SC4}} & {\centering{}{\footnotesize{}2967.38 [7.24]}} & {\centering{}{\footnotesize{}69.61 [0.125]}} & {\centering{}{\footnotesize{}0.978 [0.0044]}}\tabularnewline
\hline

{\centering{}{\footnotesize{}\textbf{SC5}}} & {\centering{}{\footnotesize{}\textbf{2969.6 [7.26]}}} & {\centering{}{\footnotesize{}\textbf{59.87 [0.126]}}} & {\centering{}{\footnotesize{}\textbf{0.973 [0.0044]}}}\tabularnewline
\hline

\end{tabular}\vspace*{0.1cm}
\caption{\label{tab:ablationTable}\scriptsize Results of the ablation experiments.}
\end{centering}\vspace*{-1cm}
\end{table}
}

\modified{\noindent\textbf{Applications. }The proposed calibration can be applied
to improve different applications such as shape-from-polarization and monocular depth
estimation. Particularly, surface reconstruction from polarization applications are often
affected by discontinuities in the estimated surfaces due to the accumulated error in the
normal's integration. Notably, these errors are linked to the error
in the estimation of the AoLP and the DoLP (due to the pixel's parameters dispersion),
which are reduced by the proposed calibration approach.

An example of robotics application is given in \cref{fig:3DreconstructionRes}, in which a
parabolic shaped piece surface is retrieved by using the SfP technique explained in
\cite{DepthReconstruction}. In this image, the highlighted error is due to the
error in the normal's vector field integration. This accumulated error makes that the point
where the end of the surface meets the beginning will not match, producing the discontinuity shown.
This reconstruction error is reduced when using the calibrated camera (highlighted area).
Although a detailed analysis of applications are out of the scope of this paper,
we are providing some reconstruction results with two recent 3D shape estimation
approaches in the supplementary material.

\begin{figure*}
\begin{centering}
\begin{tabular}{cccc}
\includegraphics[width=0.2\linewidth]{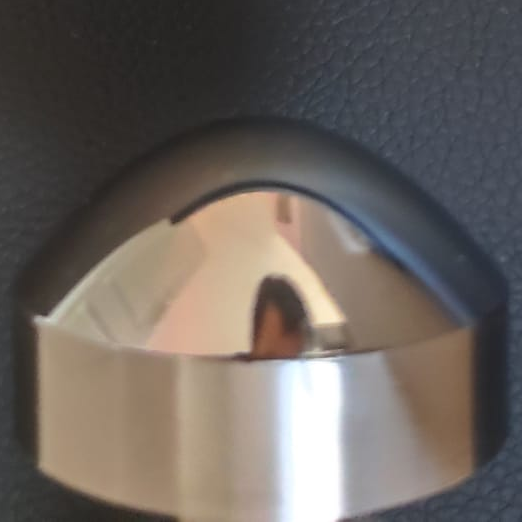} & 
\includegraphics[width=0.2\linewidth]{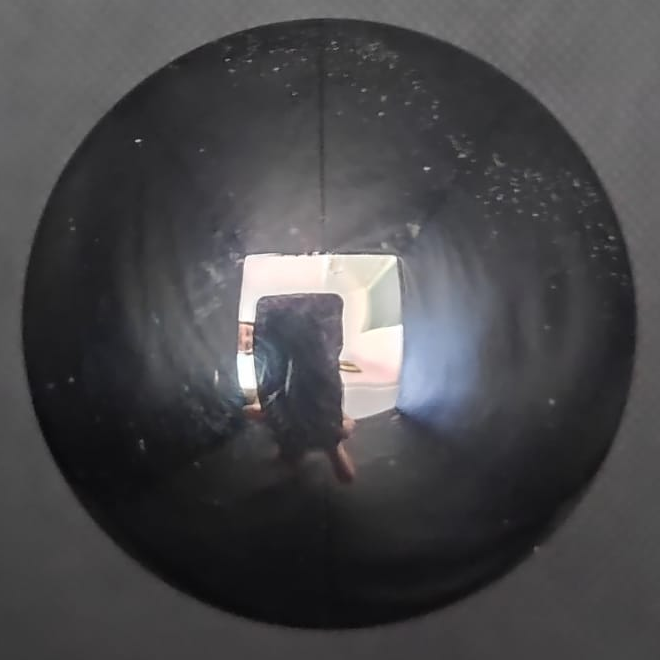} & \includegraphics[width=0.2\linewidth]{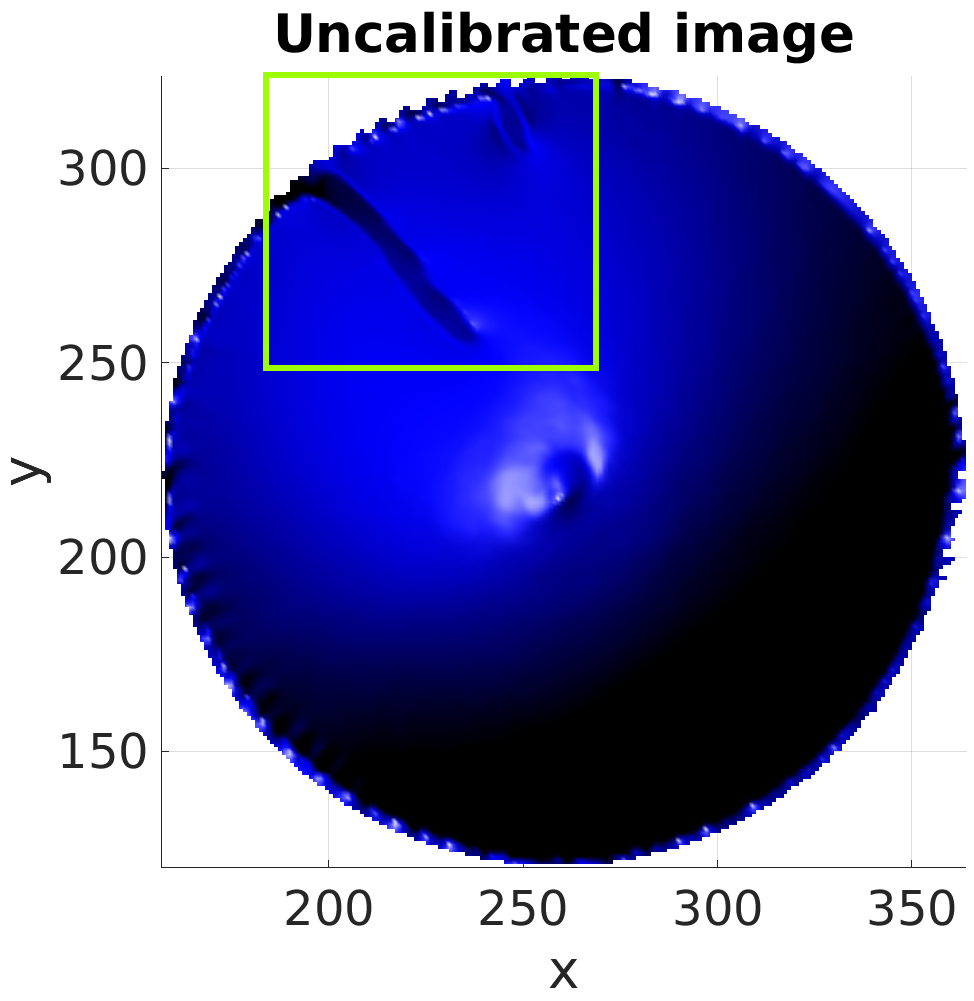} &
\includegraphics[width=0.2\linewidth]{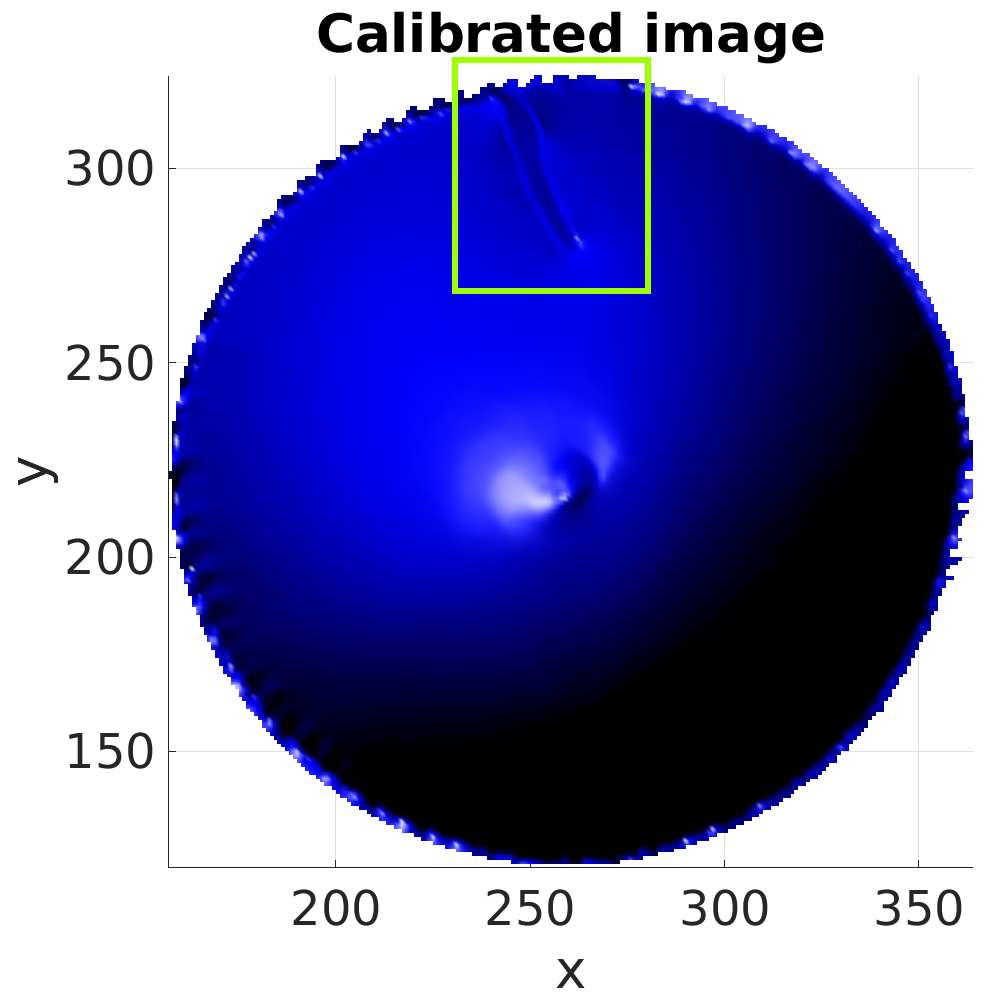} \\
(a) & (b) & (c) & (d) \\
\end{tabular}
\par\end{centering} 
\caption{\scriptsize\modified{3D reconstruction application with
    Shape-from-Polarization method~\cite{DepthReconstruction}. (a) Original
    piece -  front view. (b) Original piece - top view. (c) 3D reconstruction
    top view of the piece with the uncalibrated camera. (d) 3D reconstruction
    top view of the piece with the calibrated camera.}\vspace*{-0.7cm}}
\label{fig:3DreconstructionRes}
\end{figure*}
}

\section{\label{sec:Conclusions}Conclusions}

A calibration algorithm for RGB micro-grid
polarization cameras has been presented. The developed framework is flexible to be extended to any
amount of light samples. We show that with a minimal set-up, and without any knowledge about
the source light polarization state, the calibration problem can be solved by using
the light parameters estimators included in this work. Five image samples is the minimum
requirement to highly increase the camera estimation accuracy. The calibration light
samples must be uniform and linearly polarized, at different AoLP. Since
each color channel is considered independently from the others, the light source is not
restricted to be white, but it should have a certain value in each color frequency.
All the results in this paper have been validated with experiments,
showing that the algorithm performance is competitive to recent calibration
techniques.


{\small
\bibliographystyle{ieee_fullname}
\bibliography{references}

\begin{thebibliography}{10}\itemsep=-1pt

\bibitem{dsfp}
Yunhao Ba, Alex Gilbert, Franklin Wang, Jinfa Yang, Rui Chen, Yiqin Wang, Lei
  Yan, Boxin Shi, and Achuta Kadambi.
\newblock Deep shape from polarization.
\newblock In {\em European Conference on Computer Vision}, 2020.

\bibitem{DepthFromStereoPol}
Kai Berger, Randolph Voorhies, and Larry~H. Matthies.
\newblock Depth from stereo polarization in specular scenes for urban robotics.
\newblock In {\em IEEE International Conference on Robotics and Automation
  (ICRA)}, 2017.

\bibitem{MarcP2DApp}
Marc Blanchon, Désiré Sidibé, Olivier Morel, Ralph Seulin, Daniel Braun, and
  Fabrice Meriaudeau.
\newblock {P2D}: a self-supervised method for depth estimation from
  polarimetry.
\newblock In {\em International Conference on Pattern Recognition (ICPR)},
  2020.

\bibitem{21_Chen_calib_with_inter_fourier}
Zhenyue Chen.
\newblock Calibration method of microgrid polarimeters with image
  interpolation.
\newblock {\em Applied Optics}, 54:995--1001, 02 2015.

\bibitem{20_Gruev_Noise_in_Aop_Dop}
Yingkai Cheng, Zhongmin Zhu, Zuodong Liang, Leanne Iannucci, Spencer Lake, and
  Viktor Gruev.
\newblock Analysis of signal-to-noise ratio of angle of polarization and degree
  of polarization.
\newblock {\em OSA Continuum}, 4, 04 2021.

\bibitem{RelPosMarcPo}
Zhaopeng Cui, Viktor Larsson, and Marc Pollefeys.
\newblock Polarimetric relative pose estimation.
\newblock In {\em IEEE/CVF International Conference on Computer Vision (ICCV)},
  2019.

\bibitem{2_Non_uniform_light_calibration}
Zhichao Ding, Chunsheng Sun, Hongwei Han, Liheng Ma, and Yonggang Zhao.
\newblock Calibration method for division-of-focal-plane polarimeters using
  nonuniform light.
\newblock {\em IEEE Photonics Journal}, 12 2020.

\bibitem{23_HuangFei}
Huang Fei, Fan-Ming Li, Wei-Cong Chen, Rui Zhang, and Chao-Shuai Chen.
\newblock Calibration method for division of focal plane polarimeters.
\newblock {\em Applied Optics}, 57:4992, 06 2018.

\bibitem{polaPosePredict}
Daoyi Gao, Yitong Li, Patrick Ruhkamp, Iuliia Skobleva, Magdalena Wysocki,
  HyunJun Jung, Pengyuan Wang, Arturo Guridi, and Benjamin Busam.
\newblock Polarimetric pose prediction.
\newblock In {\em European Conference on Computer Vision (ECCV)}, October 2022.

\bibitem{9_surfacereconstruction}
Missael García, Ignacio Erausquin, Christopher Edmiston, and Viktor Gruev.
\newblock Surface normal reconstruction using circularly polarized light.
\newblock {\em Optics Express}, 23:14391, 06 2015.

\bibitem{22_Yilbert_2020}
Yilbert Gimenez, Pierre-Jean Lapray, Alban Foulonneau, and Laurent Bigué.
\newblock Calibration algorithms for polarization filter array camera: survey
  and evaluation.
\newblock {\em Journal of Electronic Imaging}, 29:1, 03 2020.

\bibitem{17_Simil_calib}
Nathan Hagen, Shuhei Shibata, and Yukitoshi Otani.
\newblock Calibration and performance assessment of microgrid polarization
  cameras.
\newblock {\em Optical Engineering}, 58:1, 02 2019.

\bibitem{FresnelEquationsBook}
E. Hecht.
\newblock {\em Optics}.
\newblock Pearson education. Addison-Wesley, 2002.

\bibitem{Ichikawa_2021_CVPR}
Tomoki Ichikawa, Matthew Purri, Ryo Kawahara, Shohei Nobuhara, Kristin Dana,
  and Ko Nishino.
\newblock Shape from sky: Polarimetric normal recovery under the sky.
\newblock In {\em IEEE/CVF Conference on Computer Vision and Pattern
  Recognition (CVPR)}, 2021.

\bibitem{6_Solar_spectropolarimetry}
Francisco Iglesias and Alex Feller.
\newblock Instrumentation for solar spectropolarimetry: State of the art and
  prospects.
\newblock {\em Optical Engineering}, 58:1, 04 2019.

\bibitem{lucid_vision_polarsens}
Lucid~Vision Labs.
\newblock Beyond conventional imaging: Sony's polarized sensor, 2022-01-30.

\bibitem{25_Connor}
Connor Lane, David Rode, and Thomas Roesgen.
\newblock Calibration of a polarization image sensor andinvestigation of
  influencing factors.
\newblock {\em Applied Optics}, 61, 10 2021.

\bibitem{14_DoT_polarimeter}
P. Marconnet, Luc Gendre, A. Foulonneau, and Laurent Bigué.
\newblock Cancellation of motion artifacts caused by a division-of-time
  polarimeter.
\newblock {\em Proc SPIE}, 8160, 09 2011.

\bibitem{13_Handy_method_Olivier}
Olivier Morel, Ralph Seulin, and David Fofi.
\newblock Handy method to calibrate division-of-amplitude polarimeters for the
  first three stokes parameters.
\newblock {\em Optics Express}, 24:13634, 06 2016.

\bibitem{10_3d_reconstruction}
Olivier Morel, Christophe Stolz, Fabrice Meriaudeau, and Patrick Gorria.
\newblock Active lighting applied to three-dimensional reconstruction of
  specular metallic surfaces by polarization imaging.
\newblock {\em Applied optics}, 45:4062--8, 07 2006.

\bibitem{16_Division_of_aperture_imaging}
Tingkui Mu, Chunmin Zhang, Qiwei Li, and Rongguang Liang.
\newblock Error analysis of single-snapshot full-stokes division-of-aperture
  imaging polarimeters.
\newblock {\em Optics Express}, 23:10822--10835, 04 2015.

\bibitem{8_water_robotics}
Samuel Powell, Roman Garnett, Justin Marshall, Charbel Rizk, and Viktor Gruev.
\newblock Bioinspired polarization vision enables underwater geolocalization.
\newblock {\em Science Advances}, 4, 04 2018.

\bibitem{18_Powell_Gruev}
S Powell and Viktor Gruev.
\newblock Calibration methods for division-of-focal-plane polarimeters.
\newblock {\em Optics express}, 21:21039--21055, 09 2013.

\bibitem{GitHubRepo}
Joaquin Rodriguez, Lew Lew-Yan-Voon, Renato Martins, and Olivier Morel.
\newblock A practical calibration method for rgb micro-grid polarimetric
  cameras: Paper supplemental material repository.
\newblock \url{https://github.com/vibot-lab/PoliCalibration}, 2022.

\bibitem{7_skin_cancer}
E. Salomatina-Motts, V. Neel, and Anna Yaroslavsky.
\newblock Multimodal polarization system for imaging skin cancer.
\newblock {\em Optics and Spectroscopy}, 107:884--890, 12 2009.

\bibitem{15_Self_calib_pola}
Yoav~Y. Schechner.
\newblock Self-calibrating imaging polarimetry.
\newblock In {\em IEEE International Conference on Computational Photography},
  2015.

\bibitem{StokesFormalism}
Robert~W. Schmieder.
\newblock Stokes-algebra formalism.
\newblock {\em J. Opt. Soc. Am.}, 59(3):297--302, Mar 1969.

\bibitem{shakeri2021polarimetric}
Moein Shakeri, Shing~Yan Loo, and Hong Zhang.
\newblock Polarimetric monocular dense mapping using relative deep depth prior,
  2021.

\bibitem{4_image_dehazing}
S. Shwartz, E. Namer, and Y.Y. Schechner.
\newblock Blind haze separation.
\newblock In {\em IEEE Computer Society Conference on Computer Vision and
  Pattern Recognition (CVPR)}, 2006.

\bibitem{DepthReconstruction}
William A.~P. Smith, Ravi Ramamoorthi, and Silvia Tozza.
\newblock Linear depth estimation from an uncalibrated, monocular polarisation
  image.
\newblock In Bastian Leibe, Jiri Matas, Nicu Sebe, and Max Welling, editors,
  {\em European Conference on Computer Vision (ECCV)}, 2016.

\bibitem{12_Overview_of_polarimetric_applications}
Frans Snik, Julia Craven-Jones, Michael Escuti, Silvano Fineschi, David
  Harrington, Antonello~De Martino, Dimitri Mawet, Jérôme Riedi, and J.~Scott
  Tyo.
\newblock {An overview of polarimetric sensing techniques and technology with
  applications to different research fields}.
\newblock In David~B. Chenault and Dennis~H. Goldstein, editors, {\em
  Polarization: Measurement, Analysis, and Remote Sensing XI}, volume 9099,
  pages 48 -- 67. International Society for Optics and Photonics, SPIE, 2014.

\bibitem{11_material_classification}
Shoji Tominaga and Akira Kimachi.
\newblock Polarization imaging for material classification.
\newblock {\em Optical Engineering - OPT ENG}, 47, 12 2008.

\bibitem{27_FullStokes_DoA}
Xingzhou Tu, Oliver~J. Spires, Xiaobo Tian, Neal Brock, Rongguang Liang, and
  Stanley Pau.
\newblock Division of amplitude rgb full-stokes camera using micro-polarizer
  arrays.
\newblock {\em Opt. Express}, 25(26), 2017.

\bibitem{24_Tyo}
J. Tyo.
\newblock Optimum linear combination strategy for an n -channel polarization-
  sensitive imaging or vision system.
\newblock {\em Journal of The Optical Society of America}, 15, 02 1998.

\bibitem{cromo}
Yannick Verdi\'e, Jifei Song, Barnab\'e Mas, Benjamin Busam, Ales Leonardis,
  and Steven McDonagh.
\newblock Cromo: Cross-modal learning for monocular depth estimation.
\newblock In {\em Proceedings of the IEEE/CVF Conference on Computer Vision and
  Pattern Recognition (CVPR)}, pages 3937--3947, June 2022.

\bibitem{lcd_calib_wang}
Zhixiang Wang, Yinqiang Zheng, and Yung-Yu Chuang.
\newblock Polarimetric camera calibration using an lcd monitor.
\newblock In {\em IEEE/CVF Conference on Computer Vision and Pattern
  Recognition (CVPR)}, 2019.

\bibitem{HDRReconsWu}
Xuesong Wu, Hong Zhang, Xiaoping Hu, Moein Shakeri, Chen Fan, and Juiwen Ting.
\newblock Hdr reconstruction based on the polarization camera.
\newblock {\em IEEE Robotics and Automation Letters}, 5(4):5113--5119, 2020.

\bibitem{5_remote_sensing}
Lei Yan, Taixia Wu, and Xueqi Wang.
\newblock Polarization remote sensing for land observation, 2018.

\bibitem{28_rt_dense_slam}
Luwei Yang, Feitong Tan, Ao Li, Zhaopeng Cui, Yasutaka Furukawa, and Ping Tan.
\newblock Polarimetric dense monocular slam.
\newblock In {\em IEEE/CVF Conference on Computer Vision and Pattern
  Recognition}, 2018.

\end{thebibliography}
}

\newpage

\begin{center}
\LARGE \sc \textbf{{Supplementary material}} \vspace*{0.5cm}
\end{center}

This is the supplementary material of our RAL 2022 submission entitled
\textbf{``A practical calibration method for RGB micro-grid polarimetric cameras"}.
This document includes additional demonstrations and experiments that might be of
interest to the readers.

\section*{Appendix A: Sine-like shape of the AoLP estimator error}
\begin{figure}[b!]
\begin{centering}
\includegraphics[width=8cm]{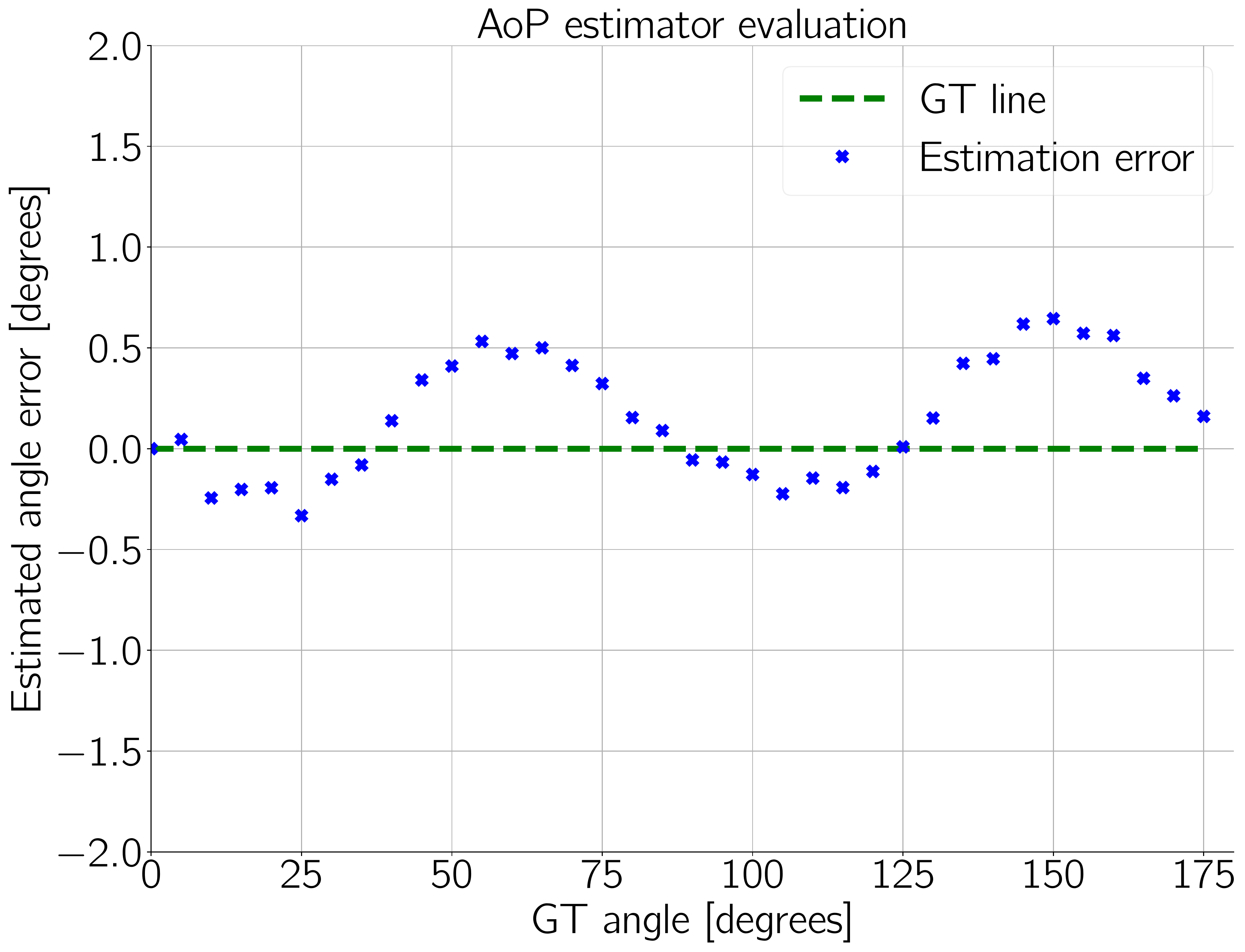}
\par\end{centering}
\caption{\label{fig:OriginalError}AoLP estimator error plot.}
\end{figure}

In this appendix, we demonstrate why the error of the AoLP estimator has a sine-like shape. This
demonstration is auxiliary to the submitted paper and the explanation was not included
in the main paper due to space restrictions. The shape shown in \cref{fig:OriginalError}
is a plot of the error in the circular average of the angle of linear polarization. The estimated
angle is calculated from the intensity measurements
of a $50\times50$ region around the center of the sensor with a 16mm lens.
For each AoLP, ten images of a uniform, linearly polarized light are captured and averaged
to reduce the influence of the noise in the estimations.

\section{Demonstration}
When a pixel and its polarization filter are not ideal, the
relationship between the Stokes vector and its measured intensity is given by
Eq. (6) in the main paper:

\begin{equation}
    I_{i}=T_{i}\left[
    \begin{array}{ccc}
        \dfrac{1}{P_{i}} & \cos\left(2\theta_{i}\right) & \sin\left(2\theta_{i}\right) \\
    \end{array}
    \right]\left[
    \begin{array}{c}
        S_{0}  \\
        S_{1}  \\
        S_{2}  \\
    \end{array}
    \right],
    \label{eq:MuellerModel}
\end{equation}
where $I_{i}$ is the measured pixel intensity, $T_{i}$ is the pixel gain,
$P_{i}$ is a factor that models the non-ideality of the pixel micro-filter,
and $\theta_{i}$ is the micro-filter orientation, for
$i=\left\{0^\circ, 45^\circ, 90^\circ, 135^\circ\right\}$.

To demonstrate the error equation in the estimated AoLP, we compute the error
in the measured Stokes components when the pixels are considered ideal.

\begin{equation}
\begin{array}{c}
    \hat{S_{1}}=I_{0} - I_{90} \\
    \hat{S_{2}}=I_{45} - I_{135} \\
\end{array}
    \label{eq:S1FirstEq}
\end{equation}

Considering the pixel model of \cref{eq:MuellerModel}, and assuming that
the micro-filter with orientation $i$ has an error $\varDelta \theta_{i}$
with respect to its ideal orientation, we obtain:

\begin{equation}
\small
\begin{array}{c}
        \hat{S_{1}}=\left[T_{0}\left[\begin{array}{ccc}\dfrac{1}{P_{0}} & \cos\left(2\varDelta \theta_{0}\right) & \sin\left(2\varDelta \theta_{0}\right)\end{array}\right]\right. - \\ \left.{T_{90}\left[\begin{array}{ccc}\dfrac{1}{P_{90}} & \cos\left(\pi + 2\varDelta \theta_{90}\right) & \sin\left(\pi + 2\varDelta \theta_{90}\right)\end{array}\right]}\right]
        \left[\begin{array}{c}
            S_{0}  \\
            S_{1}  \\
            S_{2}
        \end{array}\right]
    \end{array}
    \label{eq:S1EstimationDev}
\end{equation}

Using the sine and cosine properties, and grouping terms gives:

\begin{equation}
\hat{S_{1}}=\left[\begin{array}{ccc}
    A & B & C \\
\end{array}\right]\left[\begin{array}{c}
    S_{0}\\
    S_{1}\\
    S_{2}\\
\end{array}\right],
\label{eq:DistributedS1Error}
\end{equation}

with
\begin{equation*}
    \begin{array}{c}
    A=\dfrac{T_{0}}{P_{0}} - \dfrac{T_{90}}{P_{90}} \\
    \\
    B=T_{0}\cos\left(2\varDelta \theta_{0}\right) + T_{90}\cos\left(2\varDelta \theta_{90}\right) \\
    \\
    C=T_{0}\sin\left(2\varDelta \theta_{0}\right) + T_{90}\sin\left(2\varDelta \theta_{90}\right) \\
\end{array}
\end{equation*}

Since the angle error can be considered close to zero, then the corresponding Taylor
expansions up to order two can be used to replace the sine and cosine functions.
Moreover, by doing the matricial multiplication we obtain:

\begin{equation}
    \begin{array}{c}
        \hat{S_{1}} =AS_{0} + \left[T_{0}+T_{90} - 2T_{0}\varDelta \theta_{0}^2 - 2T_{90}\varDelta \theta_{90}^2\right]S_{1} + \\
        \\
        \left[2T_{0}\varDelta \theta_{0} + 2T_{90}\varDelta \theta_{90} \right]S_{2}. \\     
    \end{array}
    \label{eq:AfterTaylor}
\end{equation}

If the angle errors are between $\left[-10^\circ,10^\circ\right]$, the corresponding
range in radians is $\left[-0.1745,0.1745\right]$. Thus, if we square this range, we
obtain a range of values $\left[0, 0.03\right]$. The orientation errors due to
manufacturing problems have values less than $10^\circ$, therefore, the second
order variables can be ignored, i.e.:

\begin{equation}
    \hat{S_{1}} =AS_{0} + G^{'}S_{1} + K_{1}S_{2}. \\
    \label{eq:IgnoringSecondDegree}
\end{equation}
with:
\begin{equation*}
\begin{array}{c}
    G^{'}=T_{0}+T_{90} \\
    \\
    K_{1} = 2T_{0}\varDelta \theta_{0} + 2T_{90}\varDelta \theta_{90}.
\end{array}
\end{equation*}

Similarly, $\hat{S_{2}}$ can be obtained as a function of the Stokes
components.

\begin{equation}
    \hat{S_{2}} = DS_{0} - K_{2}S_{1} + G^{''}S_{2}. \\
    \label{eq:S2Estimation}
\end{equation}
where:
\begin{equation*}
\begin{array}{c}
    D=\dfrac{T_{45}}{P_{45}} - \dfrac{T_{135}}{P_{135}} \\
    \\
    G^{''}=T_{45}+T_{135} \\
    \\
    K_{2} = 2T_{45}\varDelta \theta_{45} + 2T_{135}\varDelta \theta_{135}.
\end{array}
\end{equation*}

It follows that the estimated AoLP $\hat{\alpha}$ is equal to:

\begin{equation}
\hat{\alpha}=\dfrac{1}{2}
\atan\left(\dfrac{\hat{S_{2}}}{\hat{S_{1}}}\right)=\dfrac{1}{2}\atan\left(\dfrac{DS_{0} - K_{2}S_{1} + G^{''}S_{2}}{AS_{0} + G^{'}S_{1} + K_{1}S_{2}}\right)
\label{eq:AoLPEstimation}
\end{equation}

Remembering that $S_{1}=S_{0}\rho \cos\left(2\alpha\right)$, and $S_{2}=S_{0}\rho \sin\left(2\alpha\right)$,
where $\rho$ is the degree of linear polarization, and $\alpha$ is the angle of linear polarization
of the incoming light, \cref{eq:AoLPEstimation} becomes:
\begin{equation}
    \hat{\alpha}=\dfrac{1}{2}\atan\left(\dfrac{D - K_{2}\rho \cos\left(2\alpha\right) + G^{''}\rho \sin\left(2\alpha\right)}{A + G^{'}\rho \cos\left(2\alpha\right) + K_{1}\rho \sin\left(2\alpha\right)}\right)\\
    \label{eq:FinalExpression}
\end{equation}
This equation converges to the true AoLP $\alpha$ if the pixels and the filters are
ideal, i.e., $P_{i} = 1$, $T_{i}=0.5$, and $\varDelta \theta_{i}=0$, for
$i=\left\{0, 45, 90, 135\right\}$. In a real case, slight deviations from these values
will appear. The sources of these deviations are the manufacturing process of the sensor,
and the lens added to the camera. As mentioned in the main paper, considering a small
region around the center of the sensor reduces the deviations caused by the lens.

Analyzing this equation, it is possible to conclude that:
\begin{itemize}
    \item[-] The deviations in the pixel parameters will make the other Stokes
    parameters to influence the AoLP measurement.
    \item[-] The deviations in the orientations of the micro-polarizers, denoted by
    $\varDelta \theta_{i}$, for $i=\left\{0, 45, 90, 135\right\}$ will introduce an
    error based on the value of the complementary Stokes parameter (for the measurement
    of $S_{1}$, a deviation in the orientation of the micro-polarizers will introduce
    an error based on the value of $S_{2}$, and an error based on $S_{1}$ in the
    measurement of $S_{2}$).
    \item[-] For measurements of the same light at different AoLP, the deviations in
    the non-ideality factor $P_{i}$ and the gain $T_{i}$ will produce a constant
    shift in both, numerator and denominator, of \cref{eq:FinalExpression}.
    \item[-] The values of $K_{1}$ and $K_{2}$ should be close to zero, and this can
    happen in two situations: either the pixel parameters are almost ideal and
    therefore the orientation errors are almost zero, or the pixels orientation error
    are almost the same, but in opposite directions. The second case can be understood
    by looking at the definitions of these variables. For instance,
    $K_{1} = 2T_{0}\varDelta \theta_{0} + 2T_{90}\varDelta \theta_{90}$, where $T_{0}$
    and $T_{90}$ are positive numbers, and in general, they are close to $0.5$.
    Therefore, if $\varDelta \theta_{0} \simeq -\varDelta \theta_{90}$, then $K_{1}$
    will have a very tiny value. Similarly, $K_{2}$ will be almost zero when
    $\varDelta \theta_{45}\simeq -\varDelta \theta_{135}$. Finally, it can be seen that
    the errors in the orientations can be compensated if they are in opposite directions.
\end{itemize}

In all the cases, the errors will produce sine-like functions, since they will change the
ratio of the sine to the cosine functions. Nevertheless, the effect of each parameter to
the final shape of the error is different. The error in the orientations can change only
the minimum and maximum values in the estimation error function, and the factors $T_{i}$
and $P_{i}$ can create sine shaped error functions and additionally change the position of
its extreme values.

To test this formula, the error function has been computed for two set of samples from two
different lenses. The samples to which the functions are fitted have been captured using
the RGB polarization camera with the following lenses:
\begin{itemize}
    \item[-] Lens 1: Fuji-film HF16XA-5M - F1.6/16mm
    \item[-] Lens 2: Fuji-film HF8XA-5M - F1.6/8mm
\end{itemize}

Lens 1 is the one used in the experiments in the main paper.
Additionally, both lenses have been correctly focused on the light source
used, and their F-number have been set to 3, which is higher than 2.8.
This configuration has been chosen to comply with the recommendations given by
\cite{25_Connor}.

\begin{figure}
    \centering
    \includegraphics[width=8cm]{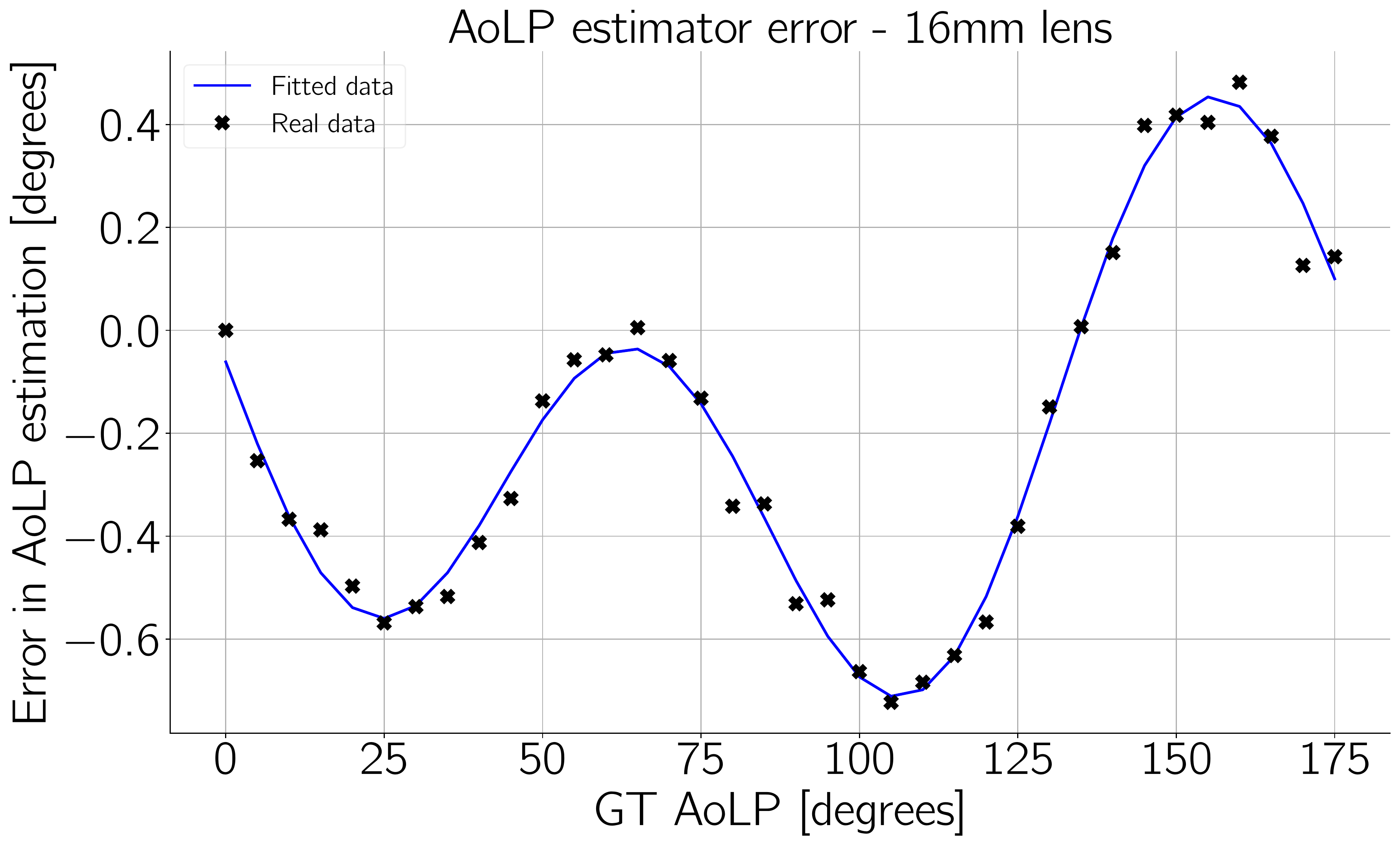}
    \caption{Error in the AoLP and curve fitted of \cref{eq:FinalExpression} for the lens 3}
    \label{fig:lens_16mm}
\end{figure}

\begin{figure}
    \centering
    \includegraphics[width=8cm]{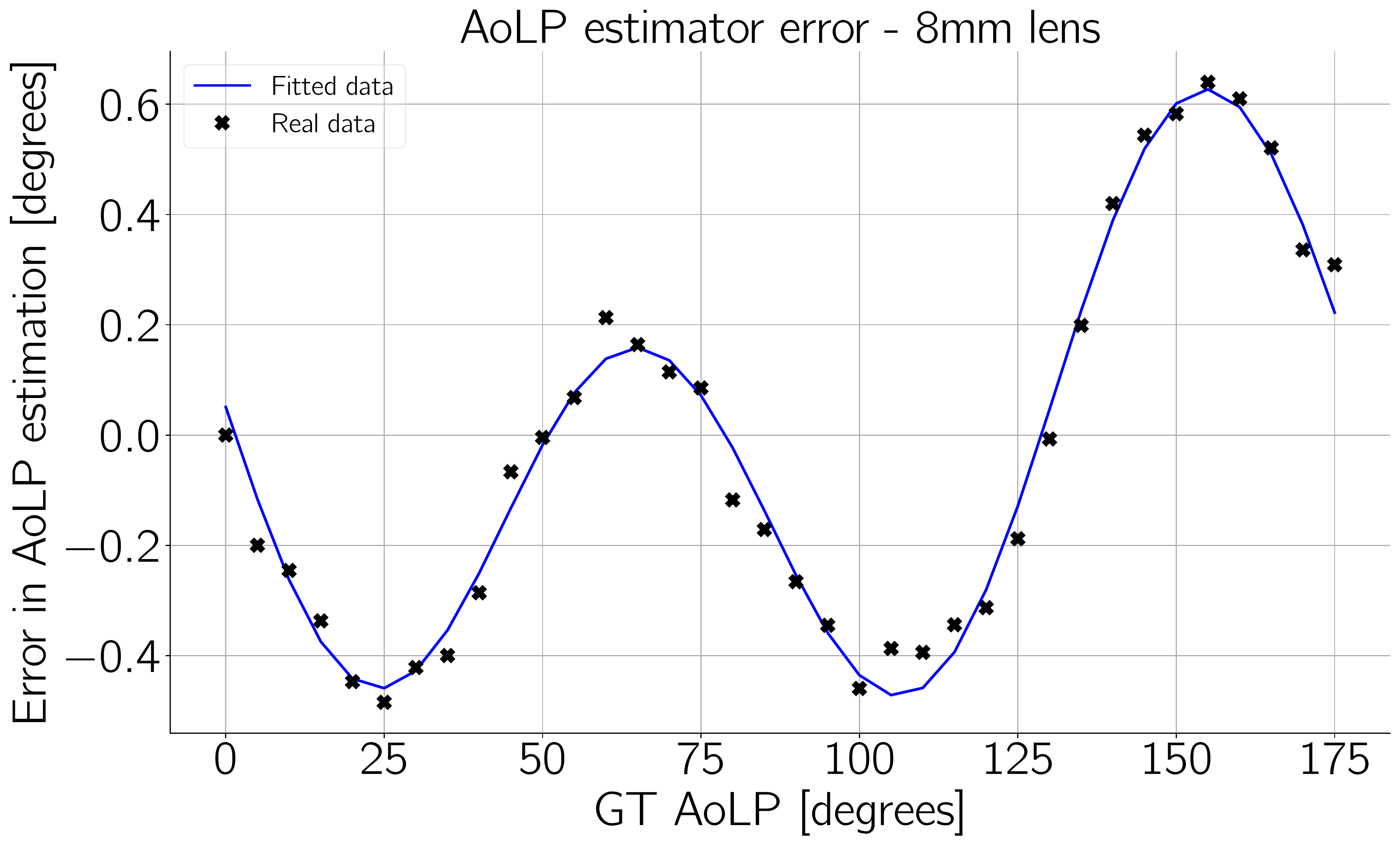}
    \caption{Error in the AoLP and curve fitted of \cref{eq:FinalExpression} for the lens 1}
    \label{fig:lens_8mm}
\end{figure}

To run this experiment, the AoLP estimator as described in the main paper has been
implemented. Then, with a uniform unpolarized light source and a rotative linear
polarization filter, a linearly polarized light is generated. The position of the
filter is changed progressively in the range $\left[0^\circ,180^\circ\right]$, with
a step of $5^\circ$. The reference angle of linear polarization of each sample has
been measured from the rotative mount of the linear filter. Additionally, the AoLP is
estimated with the implemented algorithm for each of these samples. Finally, the error
between the reference value and the estimation is computed and plotted in
\cref{fig:lens_8mm,fig:lens_16mm}.

By using a least-squares optimizer, the pixel parameters have been found for each set of
samples taken with these lenses, and the results are shown in \cref{tab:OptimizationValues}.
For creating this data, the degree of linear polarization was supposed to be $\rho=0.97$.

\begin{figure*}[t]
\begin{centering}
\begin{tabular}{cc}
\includegraphics[width=0.37\linewidth]{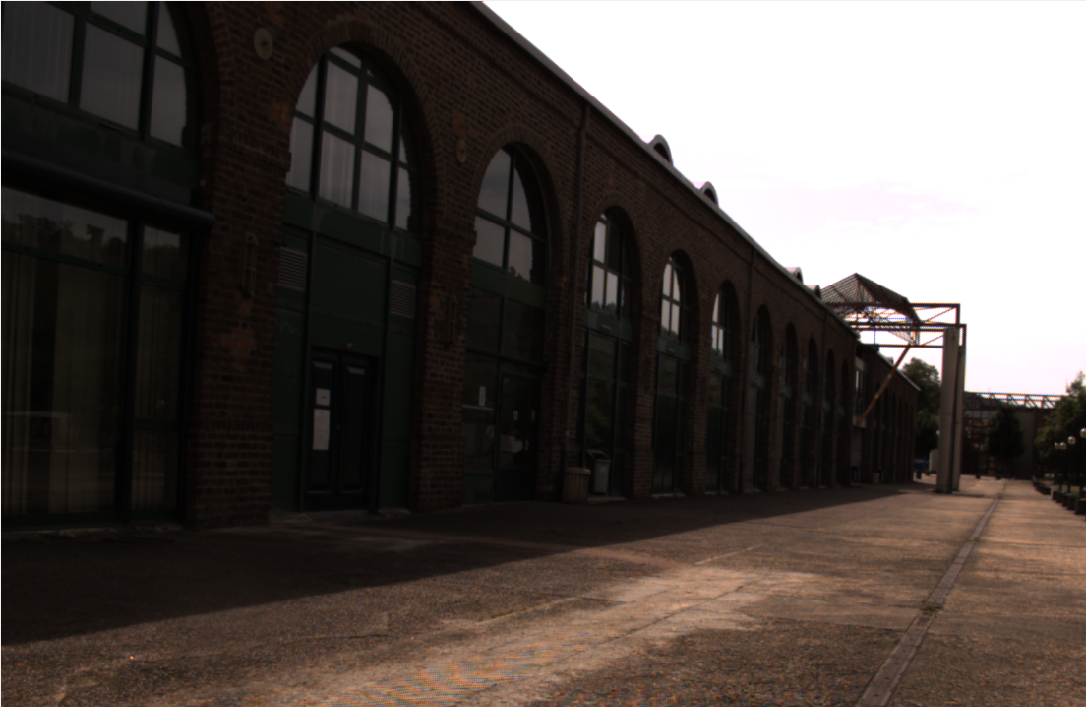} &
\includegraphics[width=0.37\linewidth]{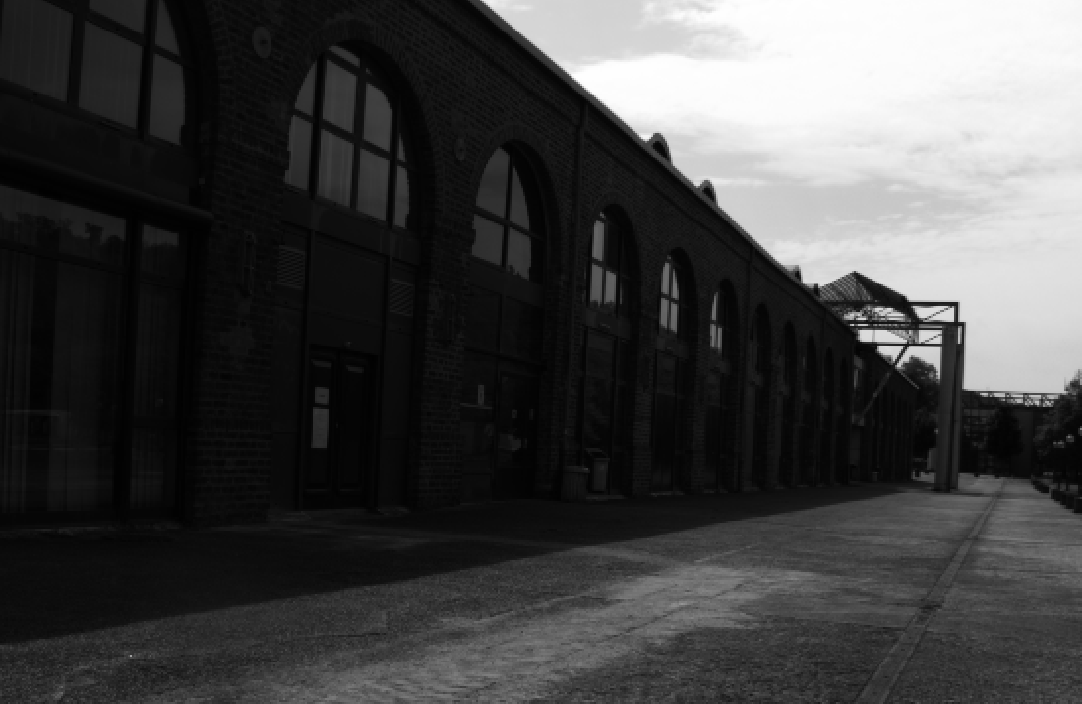} \\
(a) & (b) \\
\includegraphics[width=0.37\linewidth]{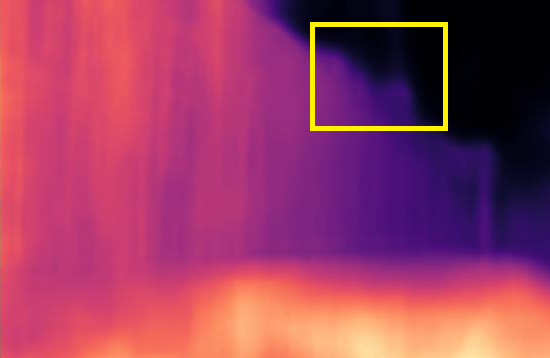}  & \includegraphics[width=0.37\linewidth]{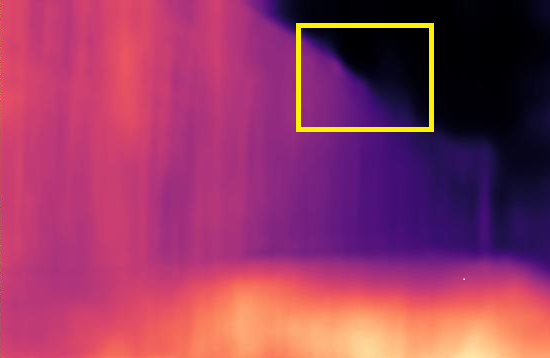} \\
(c) & (d) \\
\end{tabular}
\par\end{centering} 
\caption{Monocular disparity application example (hotter colors indicate
    larger disparity, i.e., smaller depth values). (a) Original color
    image. (b) Total intensity image (c) Disparity computed with the
    uncalibrated intensities. (d) Disparities computed with the calibrated
    intensities.}
\label{fig:MarcAppCase}
\end{figure*}

\renewcommand{\arraystretch}{1.2}
\begin{table}
    \centering
    \begin{tabular}{|c|c|c|}
    \hline
    Lens model & Parameter & $i=\left\{0,45,90,135\right\}$ \\
    \hline
    \multirow{2}{*}{$T_{i}$} & Lens 1 & $\left[0.53, 0.48, 0.49, 0.51\right]$ \\
    \cline{2-3}
    & Lens 2 & $\left[0.525, 0.542, 0.548, 0.499\right]$ \\
    \hline

    \multirow{2}{*}{$P_{i}$} & Lens 1 & $\left[1.04, 0.88, 0.97, 0.95\right]$ \\
    \cline{2-3}
    & Lens 2 & $\left[1.009, 0.992, 1.067, 0.922\right]$ \\
    \hline

    \multirow{2}{*}{$\varDelta \theta_{i}$} & Lens 1 & $\left[1.47, -1.07, -1.32, 1.54\right]$ \\
    \cline{2-3}
    & Lens 2 & $\left[0.109, 0.114, -0.192, 0.092\right]$ \\
    \hline
    \end{tabular}
    \caption{Parameters obtained by non-linear optimization for \cref{eq:FinalExpression}.}
    \label{tab:OptimizationValues}
\end{table}
\renewcommand{\arraystretch}{1}

As shown in \cref{fig:lens_8mm,fig:lens_16mm}, taking the
measurements of the AoLP from the center pixels and doing their circular average, produces
an estimation of the true AoLP with a maximum error of $0.65^\circ$. This upper limit is
valid for both lenses.

\cref{tab:OptimizationValues} shows all the pixel parameters obtained by least-squares
optimization of \cref{eq:FinalExpression} with the real data. From this table it is possible
to confirm that the effective pixel values are not far away from the ideal ones. Particularly, the
maximum orientation error is $\varDelta \theta_{0}=1.47^\circ$. Nonetheless, as explained in
the previous section, this error is compensated by the complementary pixel orientation which
is, in this case, $\varDelta \theta_{90}=-1.32^\circ$. Additionally, the values exposed in this
table show that the two lenses influence the pixel parameters. Indeed, the figures have
similar shapes, but the corresponding maximum values are not the same, and they are located at
different positions. This is because the corresponding pixel parameters have changed for
each case. It is important to highlight that the estimation error in the AoLP is limited, and small
(less than $0.65^\circ$), confirming that the initial assumption of taking central pixels
produces good estimates.

\section*{Appendix B: Applications tested with and without calibration}
Since it is important to show the benefits of calibrating the RGB polarization
camera in practical robot perception tasks, we have selected two testing
algorithms for 3D geometry estimation. One for indoor and more suitable
for small scale objects (shape-from-polarization), and one data-driven
depth estimation approach from polarization images adapted to outdoor
scenes. These algorithms are:
\begin{itemize}
    \item \textit{``Linear depth estimation from an uncalibrated, monocular
        polarisation image", ECCV, 2016} \cite{DepthReconstruction}: This paper
        proposes a shape-from-polarization algorithm with a linearization method for
        depth estimation, without any knowledge of the position of the light source.
        It is a geometric model-based approach, i.e., it does not use machine learning
        training phase for arriving to the results. We used the publicly available
        implementation in MatLab from the authors.
    \item \textit{``P2D: a self-supervised method for depth estimation from polarimetry",
        ICPR 2020} \cite{MarcP2DApp}: This paper proposes a deep learning method for
        monocular disparity estimation, using monochrome polarization images. 
        This network has been implemented considering the features given by the
        polarization state of the light. The code (in Python and PyTorch) and pre-trained
        model weights have been kindly provided by the authors of the paper upon our request.
\end{itemize}

\begin{figure}[ht]
\begin{centering}
\begin{tabular}{cc}
\includegraphics[width=0.43\linewidth]{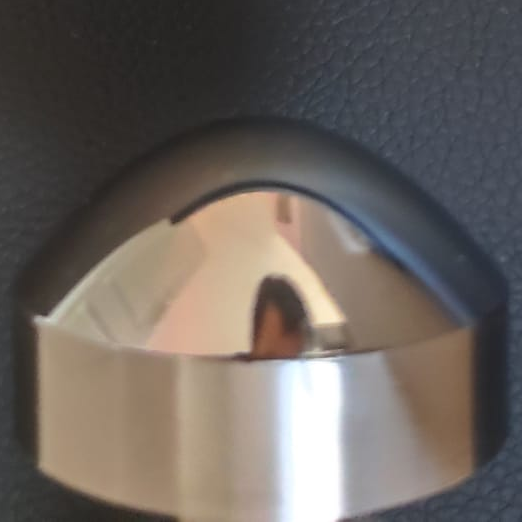} & 
\includegraphics[width=0.43\linewidth]{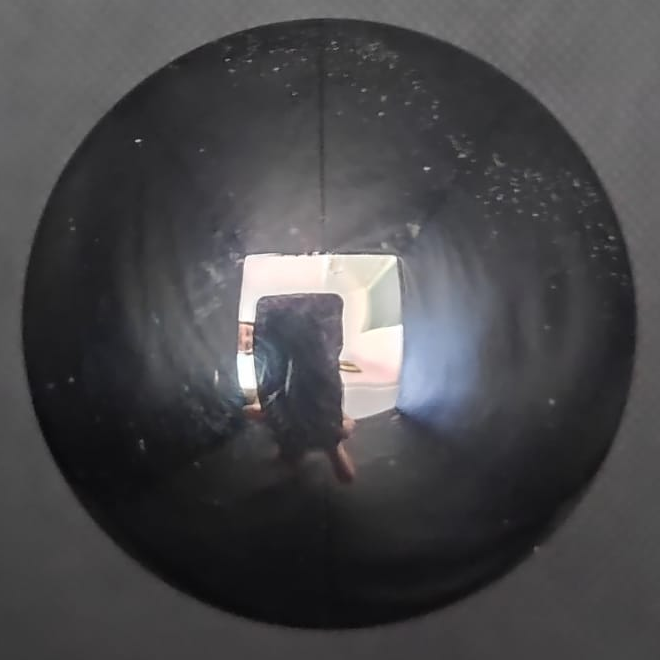} \\
(a) & (b) \\
\includegraphics[width=0.43\linewidth]{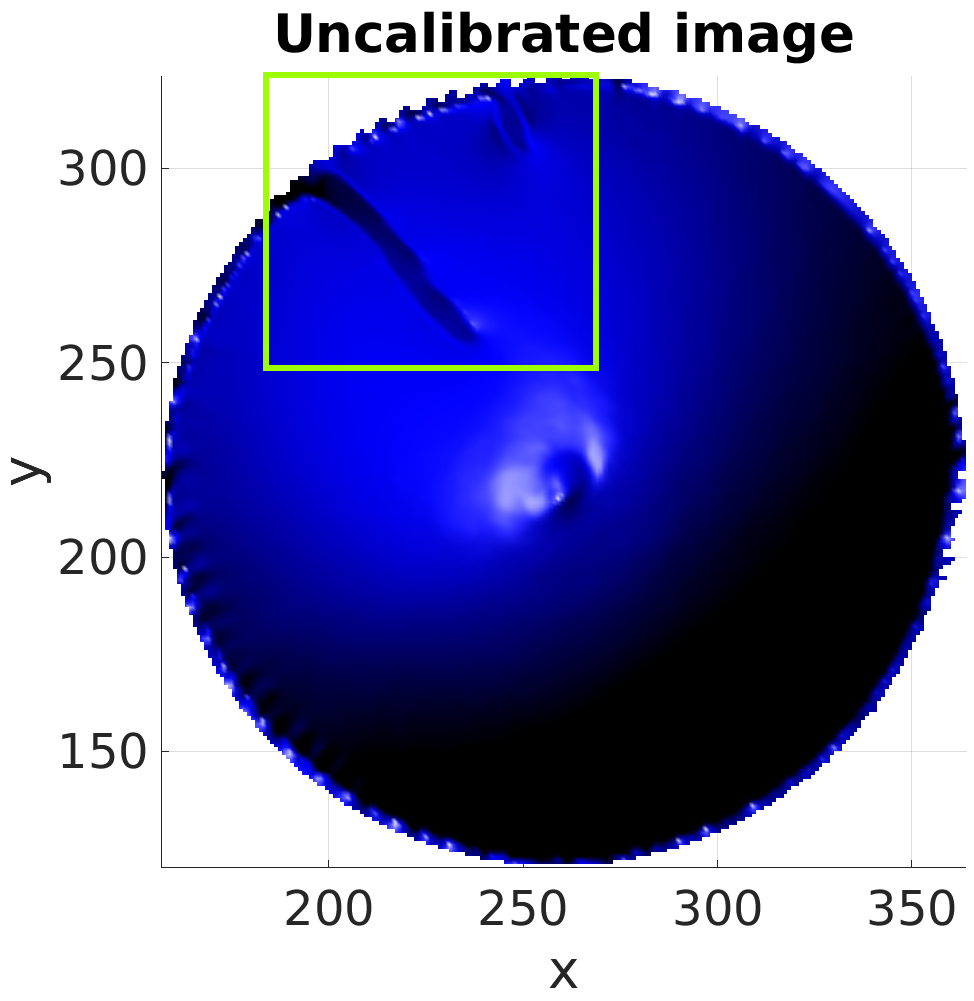} &
\includegraphics[width=0.43\linewidth]{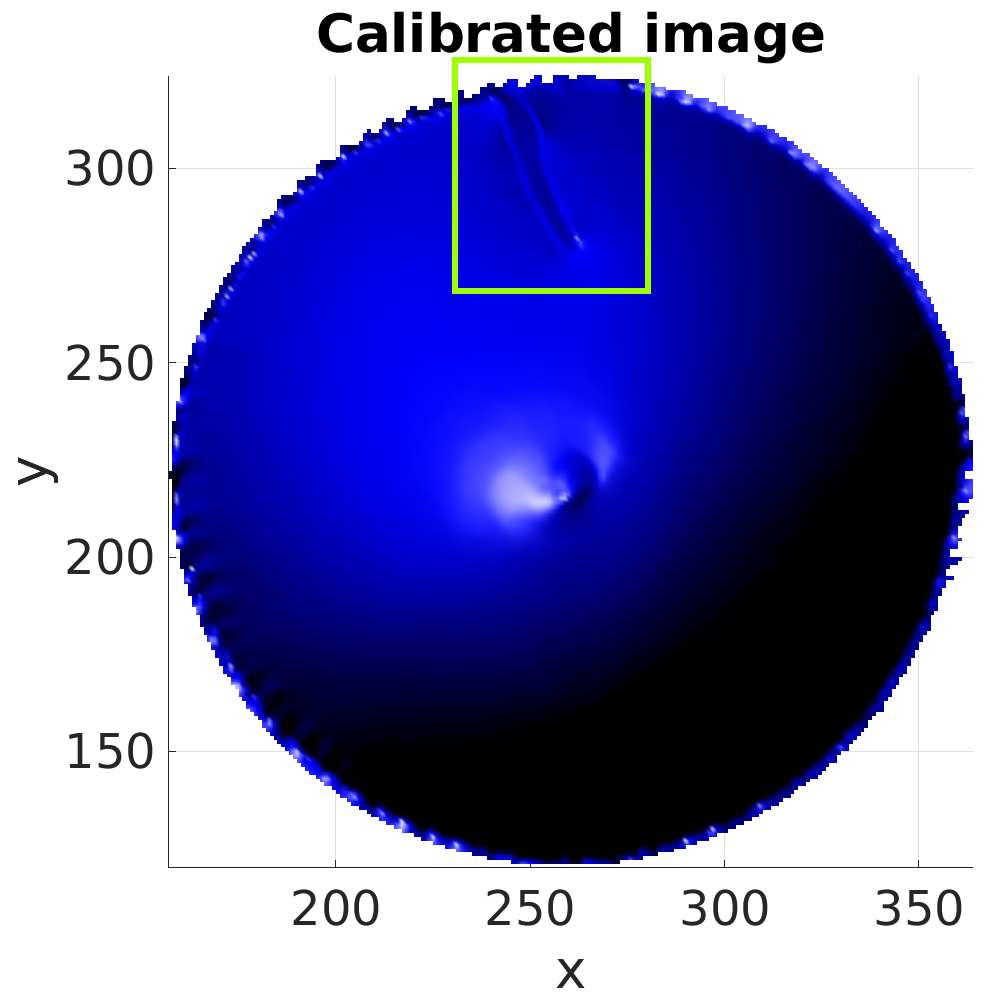} \\
(c) & (d) \\
\end{tabular}
\par\end{centering} 
\caption{3D reconstruction application result. (a) Original piece - front view.
    (b) Original piece -  top view. (c) 3D reconstruction top view of the
    piece with the uncalibrated camera. (d) 3D reconstruction top view of
    the piece with the calibrated camera.}
\label{fig:3DreconstructionRes2}
\end{figure}

The first experiment was the 3D reconstruction of a metallic, polished, parabolic object.
This object has been placed at the interior of a dome, that provides a uniform,
unpolarized, red light. The image has been captured with the camera that contains
the Sony Polarsens sensor IMX250MYR, and a lens Fujinon HF8XA-5M. The object has
been captured 1000 times, and the images have been averaged to reduce the influence
of the noise. The camera has been calibrated, and then the intensities in the average
image have been corrected with it. The reconstructed surfaces with and without
calibration are shown in \cref{fig:3DreconstructionRes2}.

The areas of interest in this figure are highlighted with a green rectangle in the
corresponding images. The corresponding images with the camera uncalibrated and calibrated are
shown in the columns (c) and (d). In the reconstruction plots, we can notice two discontinuities.
Since the surface is the result of an integration, the error in the normal's estimation
will produce that the closing points, i.e., the region where the end joins the beginning
of the surface, will not match. The error in the normal vectors is linked to the error in
the estimation of AoLP and the DoLP (due to the pixel's parameters dispersion). These
reconstruction artifacts have been reduced after calibration, with only
one discontinuity in the surface with a reduced amplitude. It is important to mention
that the same input images have been used for the two reconstructed surfaces. The
uncalibrated reconstruction is done by using the raw image from the camera (after
average), and the calibrated surface is obtained after correcting the measurements
of the raw image.
\balance
The second application corresponds to a monocular disparity estimation technique using a deep
neural network. The results before and after calibration are shown in \cref{fig:MarcAppCase}.
Again, some areas of interest are highlighted in yellow. Since the network was trained
with monochrome polarization images, the total intensity per polarization channel have been
computed and given to the network to produce the estimation. 

In this case, we can notice that after calibration the roof region is improved. In the
uncalibrated scenario, the network predicts a disparity that ascends towards the sky,
and when the image is calibrated, the disparity of this region evolves following the
border of the roof. There are two considerations that might explain this behavior: i)
The sky is polarized, so the features given by it are very distinctive from the building.
ii) Neural networks base their predictions per pixel considering its value and the ones
of the neighbors pixels. An uncalibrated setup will estimate wrong polarization
parameters due to the dispersion in the pixel parameters and due to the lens. Particularly,
as explained in the paper, the lens will produce a change in the polarization state depending
on the point of incidence of the light at the lens surface. This change will create a gradient
in the values different to the one present in the real scene. By doing the calibration, we
reduce the lens effect, and therefore making the depth estimation between objects closer to the
one in the real-scene.


\end{document}